\documentclass[conference]{IEEEtran}
\IEEEoverridecommandlockouts
\usepackage{pgfplots}
\usetikzlibrary{patterns}
\usepackage{stfloats}
\usepackage{placeins}
\usepackage{array}
\usepackage{calc}
\usepackage{array, booktabs, makecell}
\usepackage{amsmath,amssymb,amsfonts}
\usepackage{algorithmic}
\usepackage[caption=false]{subfig}
\usepackage{tikz}
\usepackage{graphicx}
\usepackage{textcomp}
\usepackage{cite}

\usepackage{hyperref}
\usepackage{cleveref}
\crefname{figure}{Figure}{Figures}
\crefname{table}{Table}{Tables}
\crefname{algorithm}{Algorithm}{Algorithms}
\crefname{equation}{Equation}{Equations}
\crefname{section}{Section}{Sections}

\usepackage{algorithm}
\usepackage{xcolor}

\pgfplotsset{compat=newest,width=8cm,height=4.8cm,
        every axis/.append style={
            tick label style={font=\fontsize{6}{6.5}\selectfont},
            label style={font=\fontsize{6}{6.5}\selectfont}
        },
        legend image code/.code={
            \draw[mark repeat=2,mark phase=2]
                plot coordinates {
                    (0cm,0cm)
                    (0.15cm,0cm)        
                    (0.3cm,0cm)
                };
        },
        major tick length=0.03cm,
        xtick align=outside,ytick align=outside,
        axis x line*=bottom,axis y line*=left,axis line style=ultra thin
}

\makeatletter 
\newcommand{\linebreakand}{%
  \end{@IEEEauthorhalign}
  \hfill\mbox{}\par
  \mbox{}\hfill\begin{@IEEEauthorhalign}
}
\makeatother 

\def\BibTeX{{\rm B\kern-.05em{\sc i\kern-.025em b}\kern-.08em
    T\kern-.1667em\lower.7ex\hbox{E}\kern-.125emX}}
\begin{document}

\title{Toward Multiple Specialty Learners for Explaining GNNs via Online Knowledge Distillation}

\author{\IEEEauthorblockN{1\textsuperscript{st} Tien-Cuong Bui}
\IEEEauthorblockA{\textit{Department of ECE} \\
\textit{Seoul National University}\\
Seoul, South Korea \\
cuongbt91@snu.ac.kr}
\and
\IEEEauthorblockN{2\textsuperscript{nd} Van-Duc Le}
\IEEEauthorblockA{\textit{Department of ECE} \\
\textit{Seoul National University}\\
Seoul, South Korea \\
levanduc@snu.ac.kr}
\and
\IEEEauthorblockN{3\textsuperscript{rd} Wen-syan Li}
\IEEEauthorblockA{\textit{Graduate School of Data Science} \\
\textit{Seoul National University}\\
Seoul, South Korea \\
wensyanli@snu.ac.kr}
\and
\IEEEauthorblockN{4\textsuperscript{th} Sang Kyun Cha}
\IEEEauthorblockA{\textit{Graduate School of Data Science} \\
\textit{Seoul National University}\\
Seoul, South Korea \\
chask@snu.ac.kr}

}

\maketitle

\begin{abstract}
Graph Neural Networks (GNNs) have become increasingly ubiquitous in numerous applications and systems, necessitating explanations of their predictions, especially when making critical decisions. However, explaining GNNs is challenging due to the complexity of graph data and model execution. Despite additional computational costs, post-hoc explanation approaches have been widely adopted due to the generality of their architectures. Intrinsically interpretable models provide instant explanations but are usually model-specific, which can only explain particular GNNs. Therefore, we propose a novel GNN explanation framework named SCALE, which is general and fast for explaining predictions. SCALE trains multiple specialty learners to explain GNNs since constructing one powerful explainer to examine attributions of interactions in input graphs is complicated. In training, a black-box GNN model guides learners based on an online knowledge distillation paradigm. In the explanation phase, explanations of predictions are provided by multiple explainers corresponding to trained learners. Specifically, edge masking and random walk with restart procedures are executed to provide structural explanations for graph-level and node-level predictions, respectively. A feature attribution module provides overall summaries and instance-level feature contributions. We compare SCALE with state-of-the-art baselines via quantitative and qualitative experiments to prove its explanation correctness and execution performance. We also conduct a series of ablation studies to understand the strengths and weaknesses of the proposed framework. 

\end{abstract}


\section{Introduction}

As graph neural networks (GNN) \cite{zhou2020graph,zhang2020deep} have become increasingly ubiquitous in real-world applications and large-scale systems, understanding the causes behind predictions is essential for assessing trust, especially regarding critical decisions. However, generating explanations for GNN is challenging due to the following reasons. Unlike tabular data, text, and grid-like data, graphs consist of multiple components, such as graph structures, node features, and edge features. Therefore, influential factors of a certain prediction can be complicated interactions among these factors. Furthermore, graph datasets are diverse, wherein each may contain a different set of components, thus making it hard to measure exact contributions. Moreover, it is not easy to use general explainable AI (XAI) tools \cite{ribeiro2016should,lundberg2017unified} for explaining GNNs due to the complexity of graphs.

\begin{figure}
    \centering
    \subfloat[Relationships of Approaches]{
        \includegraphics[width=0.46\columnwidth,trim={8.7cm 1.6cm 8.7cm 1.6cm},clip]{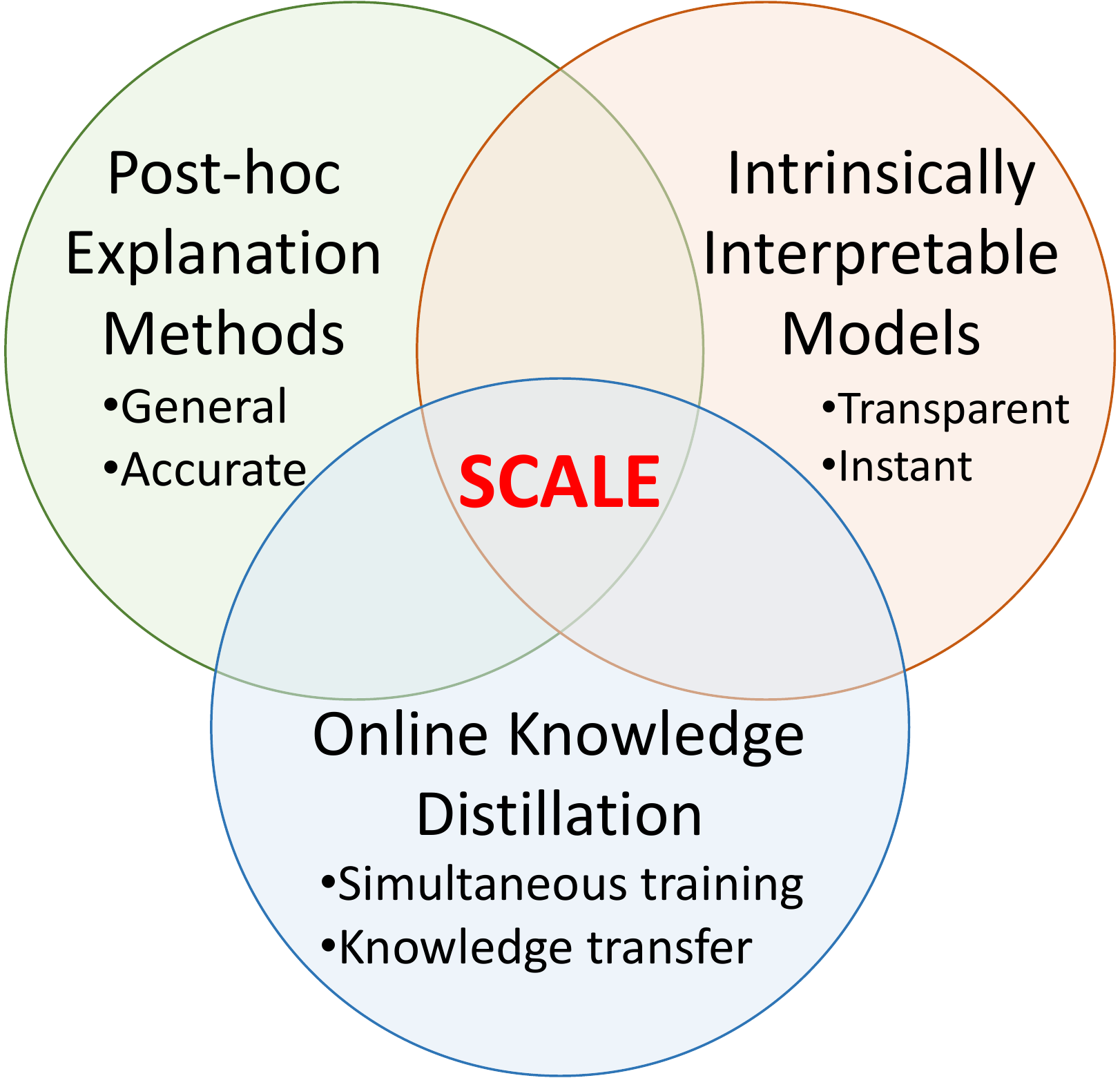}
        \label{fig:venn_SCALE}
    }
    \hspace{-0.3cm}
    \hfill
    \subfloat[Positioning Existing GNN Explanation Approaches]{
        \includegraphics[width=0.46\columnwidth,trim={7.9cm 1cm 9.2cm 1.6cm},clip]{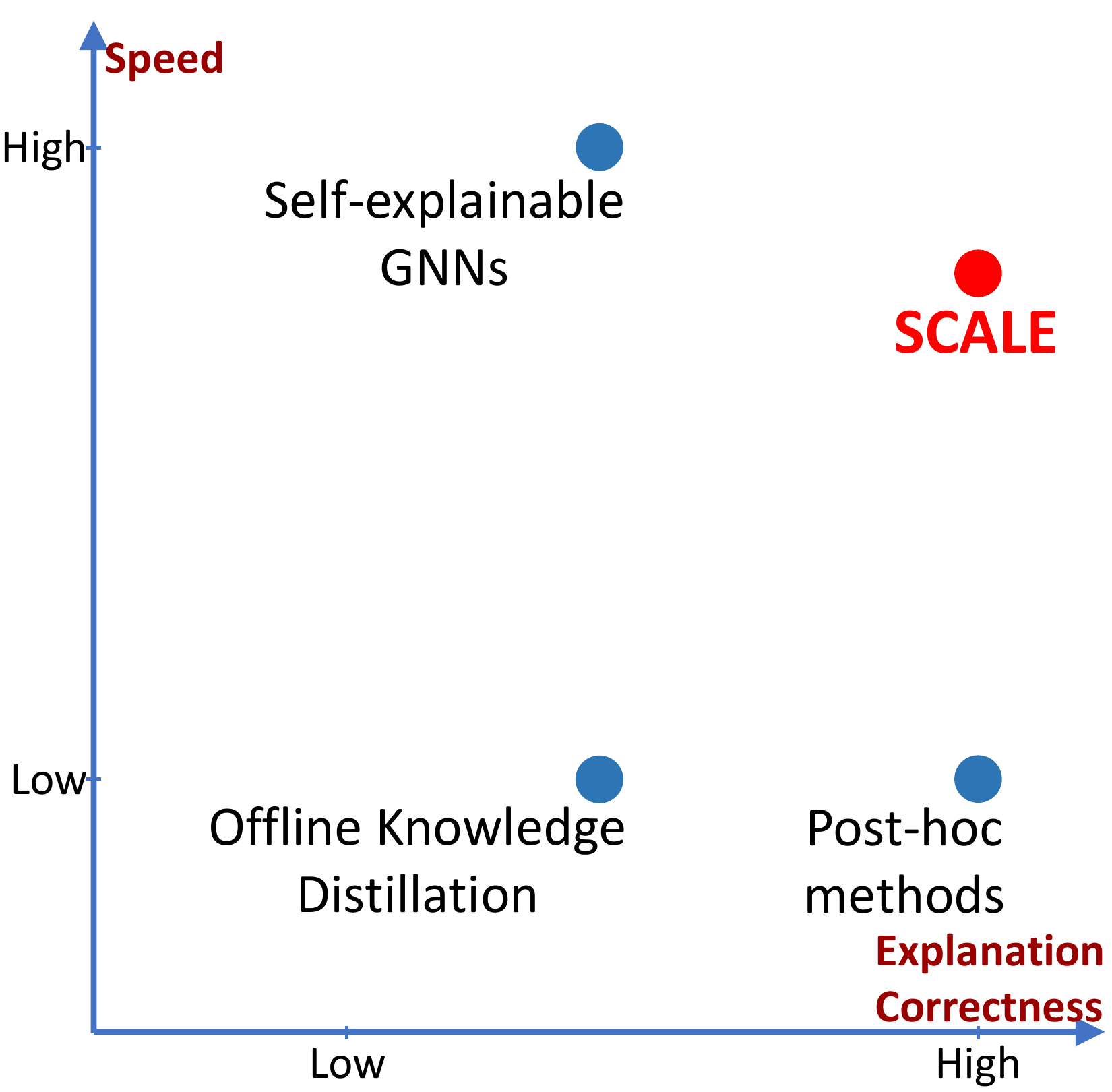}
        \label{fig:position}
    }
    \caption{SCALE inherits the best features from post-hoc explanation methods and intrinsically interpretable models for GNNs. It trains multiple specialty learners simultaneously with a black-box GNN to explain GNNs based on online knowledge distillation \cite{gou2021knowledge}. It can instantly provide accurate explanations for GNNs' predictions. }
\end{figure}

Lately, several approaches have been proposed for explaining GNNs \cite{yuan2020explainability}, as there is no one-size-fits-all solution in XAI. These methods look at the problem from different angles and provide multiple ways to explain GNNs. Most existing approaches fall under the post-hoc explanation category and focus on instance-level explanations \cite{yuan2020explainability}. Among them, perturbation methods such as GNNExplainer \cite{ying2019gnnexplainer} and PGExplainer \cite{luo2020parameterized} have been widely adopted since they introduced benchmark datasets for explanation tasks and achieved outstanding results. However, these methods need to train explanation models for target instances after training black-box GNNs resulting in additional computational resources and execution time. Unlike post-hoc methods, intrinsically interpretable (self-explainable) models such as GAT \cite{velickovic2017graph} can provide explanations for predictions instantly. Several novel self-explainable GNN architectures \cite{dai2021towards,zhang2022protgnn} have been proposed based on distance-based objective functions. However, they usually evaluate their proposed methods using citation graphs \cite{dgldata}, which lack generalization to other datasets. Moreover, similarity functions can be a computational burden in many scenarios when graphs are complex. Besides, existing methods mostly concentrate on structural explanations while overlooking feature attributions.

A promising but challenging research problem is training explainers with a black-box GNN, wherein each explainer learns to analyze a subset of interactions within an input graph. This approach enables explainers to be as general as post-hoc explanation methods and as quick as self-explainable models when making explanations. However, designing an effective training paradigm, which allows explainable models to achieve equivalent performance as the black box, is not a straightforward problem. It is because explainable models' performances can be inferior in a standalone training mode resulting in untrustworthy explanations. One solution for this problem is online knowledge distillation \cite{gou2021knowledge}, wherein a black-box GNN is regarded as a teacher, while explainable models are considered students. The teacher guides the learning processes of students via distillation losses. 

\begin{figure*}[ht]
    \centering
    \subfloat[Model Training]{
        \includegraphics[width=0.76\linewidth,trim={0.7cm 2.4cm 3.2cm 0.6cm},clip]{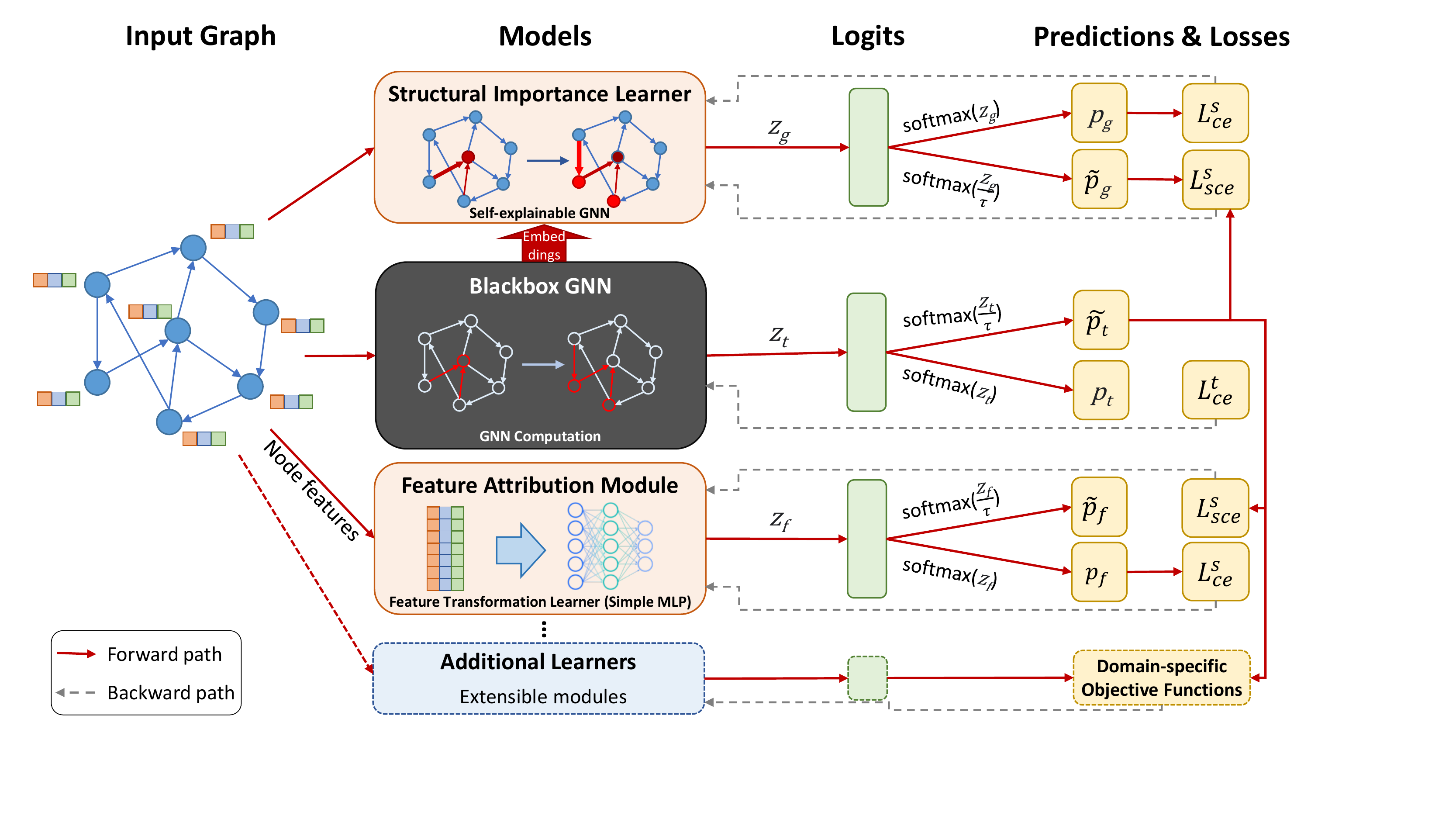}   \label{fig:overview_a}
    }
    \vspace{-0.4cm}
    \hfil
    \subfloat{
    \centering
    \resizebox{0.7\linewidth}{0.5cm}{%
    \begin{tikzpicture}
        \draw[-] (5,0) -- (20,0) node[]{};
        \end{tikzpicture}%
    }
    }
    \vspace{-0.4cm}
    \hfil
    \setcounter{subfigure}{1}
    \subfloat[Inference \& Explanation]{
        \includegraphics[width=0.6\linewidth,trim={0.7cm 4.7cm 8.1cm 1.7cm},clip]{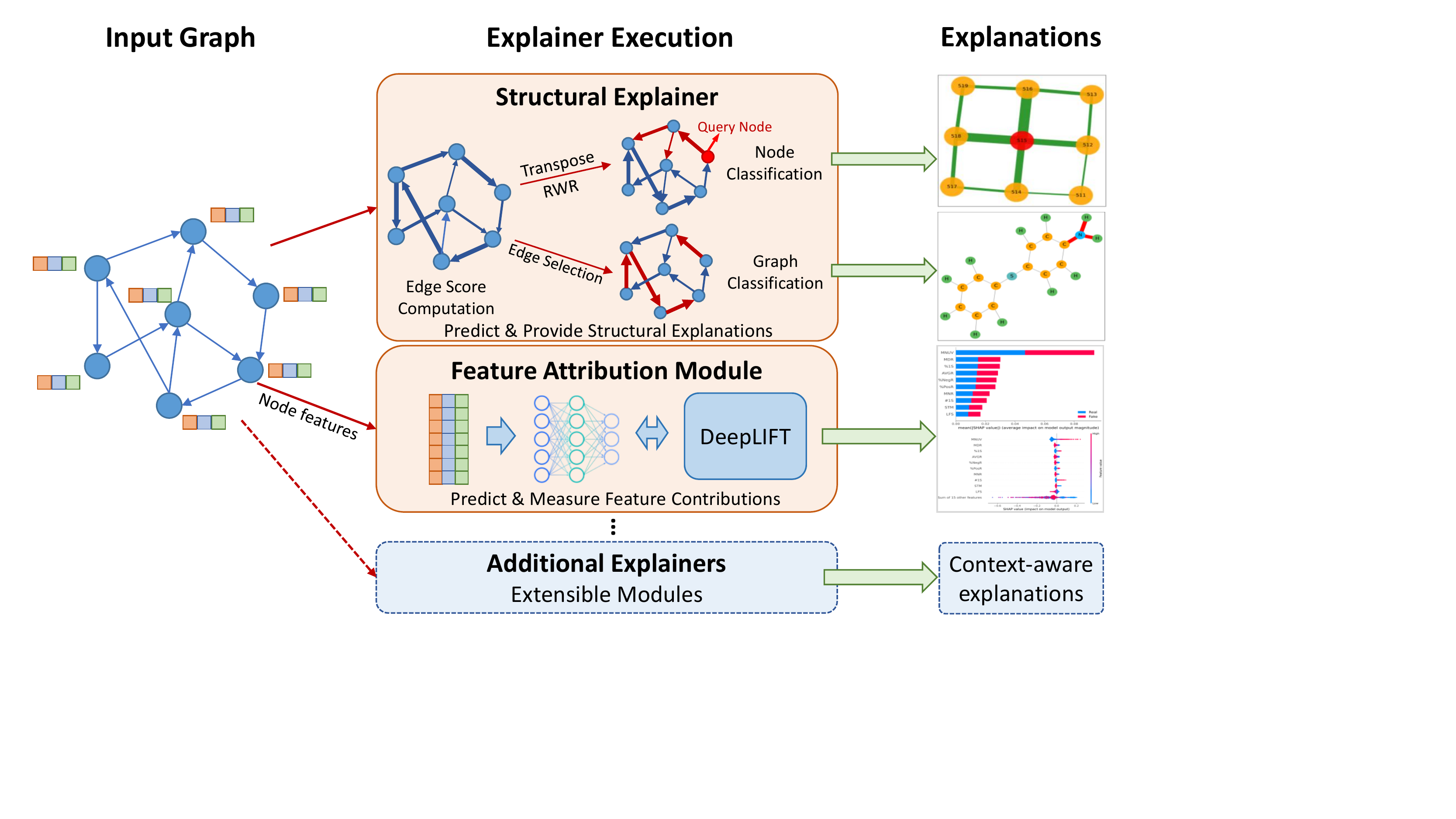}
        \label{fig:overview_b}
    }
    \caption{An illustration of SCALE Architecture. The upper figure demonstrates the training process of SCALE, while the bottom one presents how SCALE provides explanations based on trained learners. Additional learners and explainers can be implemented depending on particular requirements.}
    \label{fig:overview}
\end{figure*}
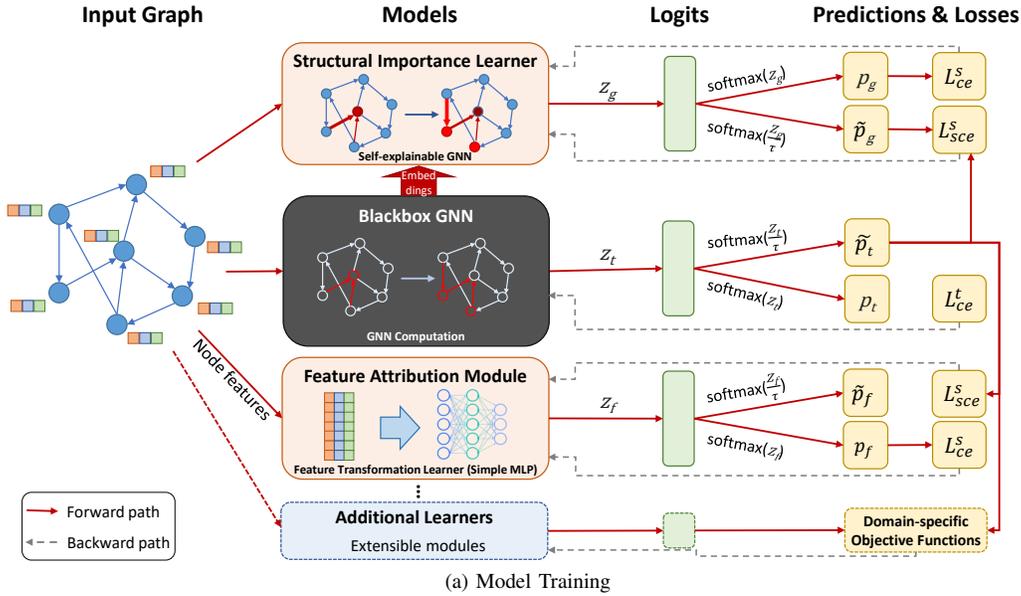
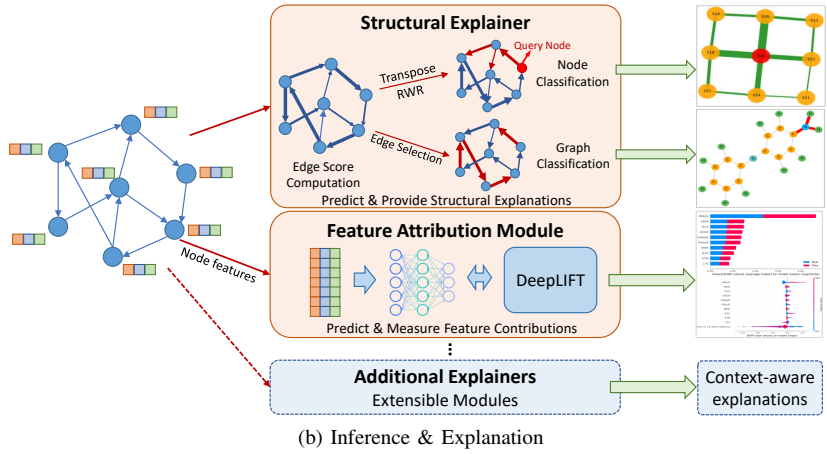

We propose SCALE, the first explanation framework training multiple \textbf{S}pe\textbf{C}i\textbf{A}lty \textbf{L}earners to \textbf{E}xplain GNNs. As depicted in \cref{fig:venn_SCALE}, SCALE combines the advantages of post-hoc explanation methods, intrinsically interpretable models, and online knowledge distillation. Following \cite{ying2019gnnexplainer, luo2020parameterized}, our explanation framework concentrates on finding important factors from graph structures, node features, and edge features that affect predictions the most. Here, we relax the complexity of the problem by discarding edge features. Since numerous interactions exist among these components, creating a single explainer to examine attributions is complicated. Therefore, SCALE trains multiple specialty learners to explain predictions in different aspects. A learner is a model that learns from a black-box GNN to capture a subset of interactions. As shown in \cref{fig:overview_a}, structural importance and feature transformation learners are trained simultaneously with the black-box GNN based on an online knowledge distillation paradigm. This paradigm enables the black-box GNN model to guide learners to achieve approximate predictive performance. 

In the explanation phase, instant explainers output predictions with explanations in different aspects and levels based on special procedures. First, an edge selection procedure eliminates unimportant edges and nodes for a graph-level prediction using learnable masks of a structural importance learner. Second, a customized version of the random walk with restart (RWR) algorithm, which has been widely applied to recommender systems and search engines \cite{brin1998anatomy,chiang2013exploring,park2017comparative,wang2020personalized}, is employed to provide structural explanations for node-level predictions.
In our scenario, a target node that needs an explanation is considered a query vertex. A random walker includes neighboring nodes and edges in the explanation as it jumps through these elements. Moreover, we execute DeepLIFT \cite{shrikumar2017learning,shapDeep19:online} on top of a trained multilayer perceptron (MLP) learner in the feature attribution module to provide instance-level feature attributions and overall attribution summaries. This library is selected since it is an effective and fast feature attribution method for deep neural networks.

Evaluating the efficiency of explanation methods is difficult since ground-truth explanations do not exist in most graph datasets. Therefore, we follow \cite{ying2019gnnexplainer,luo2020parameterized} to formulate structural explanations as binary classification tasks, wherein influential nodes and edges of predictions are included in explanations. Then, we make quantitative and qualitative comparisons between SCALE and baselines on the correctness of structural explanations and execution performance. Moreover, we evaluate the efficiency of the feature attribution module using a real-world dataset with intelligible features. Since no ground-truth information exists, we compare our observations with findings from a state-of-the-art method \cite{zhang2020gcn} for evaluation purposes. Finally, a series of ablation studies are conducted to understand algorithms in SCALE better. Through extensive experiments and analyses, we conclude that SCALE is effective in providing explaining GNNs in both explanation capability and execution performance.

Our contributions are summarized as follows:
\begin{itemize}
    \item We propose a novel explanation framework that utilizes multiple specialty learners to provide accurate explanations instantly for GNNs. 
    
    \item Our proposed framework is general and can explain any message-passing-based GNN architectures. Moreover, it provides instant explanations for input graphs.
    
    \item We are the first to use RWR for explaining node-level predictions, which can specify different contributions of neighbors to a target node.
    
    \item We conduct extensive experiments and a series of ablation studies to demonstrate SCALE's superiority in explanation capability and execution performance.
\end{itemize}

Here is the structure of the paper. \cref{related_work} describes related work. \cref{preliminary} presents background concepts. We describe the methodology in \cref{method}. Experiments are presented in \cref{exp_setups,exp_results}. We discuss detailed limitations, improvements, and possible applications in \cref{discussion}. Finally, we conclude our work in \cref{conclusion_part}.

\section{Related Work} \label{related_work}
\subsection{Post-hoc Explanation Methods}
Most existing GNN explanation methods are post-hoc approaches, which consider pre-trained models black boxes. Lately, a good survey of these methods has been provided by Yuan et al. \cite{yuan2020explainability}. Instance-level explanation receives much attention from the research community, which includes four main topics: gradient-based, decomposition, surrogate, and perturbation. While the first three approaches try to fit the existing XAI methods to the graph domain \cite{baldassarre2019explainability, schnake2020higher}, the perturbation one, first proposed by Ying et al. \cite{ying2019gnnexplainer}, is an active research area with several following research papers \cite{luo2020parameterized, schlichtkrull2020interpreting, yuan2021explainability}. The perturbation methods focus on finding important subgraphs via either edge mask learning or MCTS algorithms. However, these methods encounter overfitting issues due to the size of perturbed samples. They also do not consider feature attributions thoroughly. Moreover, post-hoc methods cannot provide instant explanations due to additional computations. Unlike them, SCALE trains specialty learners with a black-box GNN simultaneously to provide accurate explanations instantly without retraining. Besides, post-hoc explanation methods \cite{ying2019gnnexplainer,luo2020parameterized} often transform node classification problems into graph classification problems via subgraph (K-hop) sampling. This approach is suboptimal when graphs contain numerous small cycles. SCALE implements different algorithms to explain node-level and graph-level predictions.

\subsection{Self-explainable Graph Neural Networks}
Intrinsically interpretable (or self-explainable) models can overcome the performance problems of post-hoc explanation methods. Specifically, self-explainable models can provide explanations instantly based on their trained weights. Graph Attention Networks (GAT) \cite{velickovic2017graph} can be considered a self-explainable model wherein attention matrices are used for constructing explanations. GCN-LPA \cite{wang2020unifying} is another approach that integrates label propagation into GCN \cite{kipf2016semi} and replaces the normalized adjacency matrix with a learnable matrix.
Lately, several intrinsically interpretable GNN models have been proposed based on similarity functions \cite{dai2021towards,zhang2022protgnn}. 
Zhang et al. \cite{zhang2022protgnn} model the feature and label similarity in GNN execution. 
However, these methods aim to achieve high prediction accuracy on node classification datasets with the homophily property. 
Therefore, they may have inferior prediction performance on other datasets without this property, such as the ones in \cite{ying2019gnnexplainer}, leading to less trustworthy explanations.
Moreover, explanation procedures in these models are either not discussed or too straightforward for general datasets.
Even though \cite{zhang2022protgnn} can explain GNNs using prototypes extracted during training, it is significantly slow. Our proposed method aims to provide instant explanations and can be 
applied to all message-passing GNN architectures.

\subsection{Knowledge Distillation Methods for GNNs}
Knowledge distillation (KD) \cite{gou2021knowledge} was originally a model compression method \cite{hinton2015distilling}. 
Lately, many explanation methods based on it have been proposed to make student models interpretable \cite{alharbi2021learning}.
Most KD approaches for GNNs \cite{deng2021graph, joshi2021representation,yang2021extract,li2022egnn} are building small student models which can achieve or surpass the prediction accuracy of pre-trained teacher models. 
Students typically use the same datasets as the original model, as graph-free KD \cite{deng2021graph} is costly.
Even though \cite{yang2021extract,li2022egnn} propose explanation approaches for GNNs, they merely concentrate on the prediction accuracy of student models on citation graphs \cite{dgldata} or similar ones with the homophily phenomenon.
Their proposed solutions for GNN explanations are too straightforward, making it hard to achieve significant results on other datasets, such as the ones in \cite{ying2019gnnexplainer,luo2020parameterized}.
Moreover, all methods implement offline knowledge distillation paradigms that require additional computational resources.
To the best of our knowledge, we are the first to use an online knowledge distillation paradigm to train specialty learners to explain GNNs. Our proposed framework consists of multiple algorithms that explain node-level and graph-level predictions from different perspectives.

\section{Preliminaries} \label{preliminary}
An input graph $G = \{V, E, X\}$ consists of a set of vertices $V$, a set of edges $E$, and a $d$-dimensional node feature matrix $X = \{x_1,...,x_n\}, x_i \in \mathbb{R}^d$. We study the problem of explaining GNNs in node and graph classification tasks. 

\subsection{Background on Graph Neural Networks}
GNNs usually adopt message-passing paradigms which iteratively execute propagation, aggregation, and update to compute node embedding vectors. A message $m^l_{ij} = \textrm{Message}(h^{l-1}_i, h^{l-1}_j)$, where $h^{l-1}_i$ and $h^{l-1}_j$ are the representation vectors of nodes $i$ and $j$ at layer $l-1$, respectively, is sent along the edge from $j$ to $i$.
At each node, GNNs aggregate information as follows: $m^l_i = \textrm{Aggregate}({m^l_{ij}|j \in \mathcal{N}_i})$. A new representation vector at layer $l$ is computed using the following formula $h^l_i = \textrm{Update}(m^l_i, h^{l-1}_i)$. The last layer's embedding vectors $h^L$ are then used for downstream tasks such as node and graph classification.

\noindent\textbf{Node Classification} is a fundamental task in graph analytics, wherein the role is to assign correct labels to nodes in a graph. Let $f$ denote a labeling function $f: V \mapsto \{1,...,C\}$ that maps each node in $V$ to a class in $C$. In GNNs, node embedding vectors $h^L$ are fed to feed-forward networks to obtain node labels.  

\noindent\textbf{Graph Classification} refers to the problem of classifying input graphs into groups. Let $f$ denote a labeling function $f: G \mapsto \{1,...,C\}$. A pooling layer is attached on the top of node embedding vectors $h^L$ to output an embedding vector with the same dimension for an input graph. This vector is the input of feed-forward networks for classification.

\subsection{Constructing Explanations for GNNs}
A GNN's prediction $\hat{y}$ is made by a trained GNN model $f$ operating on an input graph $G_c$. Let us denote $G_c$ as the computation graph. In effect, we only need to analyze $G_c$ to explain $\hat{y}$, meaning we should focus on finding essential subgraphs and crucial node features. Here, we ignore edge features for simplicity. Formally, SCALE constructs an explanation for a prediction $\hat{y}$ as ($G_s, \Phi_x$), where $G_s$ is a subgraph of $G_c$ and $\Phi_x$ is an attribution vector of node features. Specifically, $\Phi_x = \{\phi_1, \phi_2,...,\phi_d\}$ represents the contribution of node features to the prediction $\hat{y}$. Moreover, $G_s$'s edges contain importance scores corresponding to their contribution to $\hat{y}$. 

\section{Methodology} \label{method}
\subsection{Framework Overview}

We develop SCALE based on observations of existing GNN explanation methods. First, most post-hoc explanation methods aim at finding important factors that affect particular predictions from graph structures and node features. Therefore, they design explainable models based on prior knowledge of a pre-trained GNN, such as node embeddings or predicted labels. However, post-hoc computations hinder them from providing instant explanations. Even though intrinsically interpretable models can provide instant explanations without retraining, they are mainly based on model-specific approaches evaluated with simple datasets, thus lacking generalization to broader scenarios. Moreover, all methods mainly concentrate on structural explanations while overlooking feature attributions.

SCALE is a model-agnostic framework that provides instant explanations without retraining explainers. It is like an adapter with various transformation engines that convert message-passing-based GNN architectures into explainable versions. Specifically, it trains multiple specialty learners simultaneously with a black-box GNN to explain different aspects of an input graph based on an online knowledge distillation paradigm. After training, multiple instant explainers operate on trained learners to provide predictions and explanations simultaneously. For clarity, here are the brief descriptions of learners and explainers.

\begin{itemize}
    \item \textbf{Learner}: a learner is a model guided by a black-box GNN model in training to capture special interactions in the GNN.
    \item \textbf{Explainer}: an explainer is a procedure that provides explanations for predictions based on a trained learner.
\end{itemize}

In the following subsections, we describe the training paradigm and present procedures for providing explanations. Then, we analyze the computational complexity of SCALE's internal engines in training and inference (including explanation) to demonstrate its capabilities and limitations. 

\subsection{Online Knowledge Distillation Paradigm}
SCALE trains multiple learners to capture specific aspects of a black-box model, wherein each one is then used to construct explainers. For simplicity, we implement two learners for structural explanations and node feature attributions. As presented in \cref{fig:overview_a}, structural importance and feature transformation learners are trained together with a black-box GNN based on an online knowledge distillation paradigm, which is customized from \cite{hinton2015distilling}. Specifically, learners are students in the training scheme, while a black-box GNN model serves as the teacher to guide student learning. Moreover, SCALE separates students' computational paths so that they cannot affect the teacher's performance during training. The teacher model is trained with a cross-entropy loss function as follows:
\begin{equation}
    \mathcal{L}_{ce}^t = -\frac{1}{N} \sum_{i=1}^N y_i \cdot \textrm{log}(softmax(z^t_i)),
    \label{eq:cross_entropy}
\end{equation}
where $z^t_i$ is the $i^{th}$ output vector, and $y_i$ is the $i^{th}$ one-hot label vector. $N$ denotes the number of training samples. Student models are trained with a joint objective function as follows:
\begin{equation}
    \mathcal{L}^s = \mathcal{L}_{ce}^s + \lambda \mathcal{L}_{sce}^s,
    \label{combined_student_loss}
\end{equation}
where $\mathcal{L}_{ce}^s$ is a cross-entropy loss function, and $\mathcal{L}_{sce}^s$ is a soft cross-entropy objective function based on soft targets derived from predictive distributions. $\lambda$ is a balancing factor regulating the amount of distilled information used in a student model. Here, the soft cross-entropy loss function is used instead of the KL divergence loss function due to practical performance achievements in experiments. $\mathcal{L}_{ce}^s$ and $\mathcal{L}_{sce}^s$ are as follows:
\begin{equation}
    \begin{aligned}
        \mathcal{L}_{ce}^s &= -\frac{1}{N} \sum_{i=1}^{N} y_i \cdot \textrm{log}(softmax(z^s_i)), \\
        \mathcal{L}_{sce}^s &= -\frac{1}{N} \sum_{i=1}^N softmax(\frac{z^t_i}{\tau}) \cdot \textrm{log}(softmax(\frac{z^s_i}{\tau})),
    \end{aligned}
\end{equation}
where $z^s_i$ is the $i^{th}$ output of the student model's prediction layer, and $\tau$ is the temperature term scaling information from corresponding models.

Batch normalization \cite{ioffe2015batch} is applied to multiple fully-connected layers in student networks to mitigate the changes in hidden outputs. Without batch normalization, the prediction accuracy of students is unstable in experiments leading to untrustworthy explanations. \cref{on_the_fly_algorithm} presents a joint training procedure of a black-box model and learners using the objective functions above.

\begin{algorithm}[ht]
  \caption{Joint Training Procedure}
  \begingroup
      \raggedright
      \textbf{Input}: Training dataset $\mathbb{T}$, Number of epochs $T$ \\
      \textbf{Output}: $f, g$ \\
  \endgroup
  \begin{algorithmic}[1]
  \FOR{$ epoch = 0 \rightarrow T$}
    \STATE{$f$ = train($\mathbb{T}$, $\mathcal{L}_{ce}^t$)} \COMMENT{Train teacher $f$}
    \STATE{$g$ = distill($\mathbb{T}$, $f$, $\mathcal{L}^s$))} \COMMENT{Distill knowledge to students}
  \ENDFOR
  \end{algorithmic}
\label{on_the_fly_algorithm}
\end{algorithm}

\subsection{Providing Structural Explanations}
Structural explanations are the main focus of most existing GNN explanation methods \cite{yuan2020explainability}. They aim to specify important nodes and edges to a prediction. For simplicity, we select GCN \cite{kipf2016semi} as the black-box GNN model and describe how to design structural importance learners (self-explainable GNNs) for node and graph classification problems. The simplest form of a layer-wise propagation rule is as follows:

\begin{equation}
    f(H^l, A) = \sigma(A H^l W^l),
    \label{basic_gcn}
\end{equation}
where $A$ is the adjacency matrix of an input graph, $H^l$ is a representation matrix of nodes, $W^l$ is a trainable weight matrix, and $\sigma$ is a non-linear function. 

\noindent\textbf{Graph Classification.}
A self-explainable GNN is constructed by adding a mask matrix $M$ to the propagation formula \cref{basic_gcn} as follows:
\begin{equation}
    f(H^l, A, M) = \sigma((A \odot M)H^l W^l).
    \label{graph_classification_mask}
\end{equation}

Inspired by \cite{luo2020parameterized}, the mask matrix is initialized via an MLP network in which inputs are edge embedding vectors constructed by concatenating embedding vectors of source and target nodes taken from the black-box GNN model. Specifically, each element of $M$ is computed as follows:

\begin{equation}
    m_{ij} = sigmoid(\textrm{MLP}([h_i, h_j])),
\end{equation}

\noindent where $h_i$ and $h_j$ are representation vectors of the source and target nodes of an edge. A batch normalization layer follows each layer in the MLP model to mitigate the covariate shift problem caused by the weight updates of the black-box model. The last layer outputs probabilities of whether edges are selected or not. A self-explainable GNN model is trained with the black-box GNN using \cref{on_the_fly_algorithm}, as illustrated in \cref{fig:overview_a}. In the inference and explanation phase, $M$ is used to provide a structural explanation for a particular prediction of an input graph, wherein each element tells us whether the corresponding edge is important to the prediction or not.

\noindent\textbf{Node Classification.}
Most existing methods use the same approach as graph classification problems to explain a node's prediction. Specifically, a subgraph including the target node is extracted and fed to an explainable model to obtain a structural explanation. Even though this approach can select and filter out uninfluential edges and neighbors, it cannot specify the exact contributions of these factors to a prediction. 

We follow the intuition that influential nodes are usually close neighbors. Therefore, if a neighbor is important to the target node's prediction, the edge between them should receive a high importance score. RWR is an efficient method for calculating relevance scores between two nodes. Inspired by its efficiency in recommender systems \cite{chiang2013exploring,park2017comparative,wang2020personalized}, we implement a customized version to provide structural explanations for node-level predictions. Let us recall the formula of the algorithm as follows:
\begin{equation}
    r_{t+1} = (1 - d)r_0 + d \hat{\mathcal{A}}_c r_t,
\end{equation}
where $\hat{\mathcal{A}}_c$ is a column-wise normalized transition matrix, $r_t$ is a probability distribution vector at time $t$, $r_0$ is the initial probability distribution vector, and $d$ is the probability that random walkers jump to a new state. 

A target node can be considered a query node in the RWR algorithm for explaining node-level predictions, wherein its corresponding element in $r_0$ is initialized as 1, while others are set to 0. The transition matrix $\hat{\mathcal{A}}_c$ is the transposed version of a trainable adjacency matrix $\hat{\mathcal{A}}$ taken from a structural importance learner, as presented in \cref{node_self_exp}. 
\begin{equation}
    f(H^l, \hat{\mathcal{A}}) = \sigma(\hat{\mathcal{A}} H^l W^l)
    \label{node_self_exp}
\end{equation}
\noindent If two nodes $i$ and $j$ have an edge, the corresponding element in $\hat{\mathcal{A}}$ is computed as follows:
\begin{equation}
    \hat{a}_{ij} = softmax(\textrm{MLP}([h_i, h_j])),
\end{equation}
where $h_i$ and $h_j$ are representation vectors of nodes $i$ and $j$ taken from the black-box GNN, respectively. The softmax function is executed on each row of matrix $\hat{\mathcal{A}}$.
A converged vector $r$ tells us the importance of nodes in the input graph to the target node's prediction. To visualize an explanation, we sample an important subgraph, including the target node and top $k$ nodes selected by ordering probabilities in $r$. Note that these $k$ vertices can include neighbors in multiple hops. The complete procedure is described in \cref{querying_rwr}.

\begin{algorithm}[ht]
  \caption{Querying Influential Vertices and Edges of a Node's Prediction}
  \begingroup
      \raggedright
      \textbf{Input}: A target node $v$ 
            \\\hspace{0.92cm} Pre-trained Matrix $\hat{\mathcal{A}}$, 
            \\\hspace{0.92cm} Num. of iteration $T$, 
            \\\hspace{0.92cm} Jumping probability $d$,
            \\\hspace{0.92cm} Num. of Response Nodes $k$ \\
      \textbf{Output}: An explanation of $v$'s prediction \\
  \endgroup
  \begin{algorithmic}[1]
    \STATE{$\hat{\mathcal{A}}_c$ = transpose($\hat{\mathcal{A}}$)} 
    \STATE{$\mathcal{P}_V$ = RWR($v, \hat{\mathcal{A}}_c, T, d$)} \COMMENT{Node scores}
    \STATE{$\mathcal{P}_E$ = $\textrm{diag}(\mathcal{P}_V) \cdot \hat{\mathcal{A}}$} \COMMENT{Edge scores}
    \STATE{$\mathcal{R}_V$ = top\_k($\mathcal{P}_V, k$)} \COMMENT{Top $k$ neighbors}
    \STATE{visualize($\mathcal{P}_V, \mathcal{P}_E, \mathcal{R}_V$)} \COMMENT{Visualize the explanation}
  \end{algorithmic}
\label{querying_rwr}
\end{algorithm}

\subsection{Feature Attribution Analysis}
In many real-world scenarios, meaningful node features significantly benefit machine learning models. Masking techniques proposed by Ying et al. \cite{ying2019gnnexplainer} cannot clarify the exact contributions of features to predictions. Due to the complexity of message-passing patterns, it is difficult to utilize general XAI tools \cite{ribeiro2016should, lundberg2017unified} to measure feature attributions. Moreover, we cannot examine feature attributions using transformation matrices of a pre-trained GNN since the model accuracy significantly drops when the graph structure is ignored, causing inconsistent results.

To solve the problems above, we construct a feature attribution module to output attributions for a particular prediction and a contribution summary for a group of predictions. This module consists of a feature transformation learner, which is a simple MLP model trained with the black-box GNN using \cref{on_the_fly_algorithm}. In training, \cref{on_the_fly_algorithm} allows the black-box teacher to guide the MLP student to achieve approximate predictive performance. DeepLIFT \cite{shrikumar2017learning,shapDeep19:online} is then executed on top of the MLP model in the explanation phase to produce feature attributions for predictions. In reality, several methods, such as \cite{tsang2020does}, can be integrated into SCALE since examining feature attributions is of great interest to the XAI research community. DeepLIFT is selected since it is an effective method for decomposing feature contributions in deep learning models and is very fast to compute.

\subsection{Computational Complexity}
\noindent\textbf{Training.} The computational cost grows linearly with the number of learners. Let $N$ be the total learners, and the training time is approximately $N+1$ times the time used for training a black-box GNN. Similarly, we need approximately $N+1$ times GPU memory space for keeping models. We discuss acceleration methods for this limitation in \cref{discussion}.   
Next, the number of parameters in a self-explainable GNN equals the number of weights in the black-box GNN plus the number of parameters of an MLP used for computing edge weights. Similarly, the computational cost of the feature attribution module depends on the size of the student MLP model. Graphs are loaded only once into the GPU and shared among models to reduce memory consumption.

\noindent\textbf{Inference \& Explanation.} We can use either the black-box model or explainable modules for inference since they achieve approximately the same accuracy. An explainable module provides both a prediction and an explanation together. Therefore, the computation cost is the cost of executing the predictive model plus the cost of making an explanation. For instance, the running time of \cref{querying_rwr} is counted in constructing structural explanations for node-level predictions. Similarly, finding feature importances for a particular instance requires computation time for DeepLIFT execution.

\section{Experimental Settings}\label{exp_setups}

\subsection{Objectives}
Our main focus was to present SCALE's correctness and execution efficiency. First, we compared SCALE with selected baselines to differentiate our framework from them via quantitative comparisons. We aimed to show that SCALE is better than post-hoc explanation methods in both aspects and is superior to intrinsically interpretable models on explanation correctness. Second, we compared our framework with two state-of-the-art post-hoc explanation methods \cite{ying2019gnnexplainer,luo2020parameterized} in qualitative aspects to highlight the quality of explanations provided by SCALE. Third, we assessed the feature attribution module by confirming our observations on its provided results for the Amazon dataset with findings from a data mining-based method \cite{zhang2020gcn}. SCALE provided more detailed information in structural explanations and multi-level feature contributions compared to baselines. Finally, several ablation studies were conducted to verify our framework's efficiency from different aspects. 

\subsection{Datasets}

We conducted experiments with five node classification and two graph classification datasets, as presented in \cref{tab:dataset}. Except for the Amazon dataset, others have been commonly used to perform functionally-grounded evaluations \cite{doshi2017towards} of GNN explanation methods. 

\begin{table}[ht]
    \centering
    \begin{tabular}{c|c|c|c|c|c}
        \hline
              & \#graphs & \#nodes & \#edges & \#feat. & \#labels \\
        \hline
        BA-Shapes     & 1    & 700   & - & 10 & 4 \\
        BA-Community  & 1    & 1400  & - & 10 & 8 \\
        Tree-Cycle    & 1    & 871   & - & 10 & 2 \\
        Tree-Grid     & 1    & 1231  & - & 10 & 2 \\
        Amazon        & 1    & 11.9K & 351.2K & 25 & 2\\
        \hline
        BA-2motifs    & 1K   & 25K    & 51.4K & 10 & 2 \\
        Mutag         & 4.3K & 131.5K & 266.9K & 14 & 2 \\
        \hline
        
    \end{tabular}
    \vspace{0.1cm}
    \caption{Dataset Information. K means a thousand. The number of edges in the first four datasets varies in experiments.}
    
    \label{tab:dataset}
\end{table}

\noindent\textbf{Node Classification.} Four synthetic graphs with ground-truth explanations, provided by \cite{ying2019gnnexplainer}, were used to evaluate the correctness and quality of structural explanations provided by methods.
At each synthesis, the number of nodes was kept constant while the number of edges was varied. Specifically, BA-Shapes (BA-S) was constructed by attaching 80 five-node houses to a 300-node BA graph. BA-Community (BA-C) was created by joining two BA-Shapes graphs. Similarly, Tree-Cycle (Tree-C) and Tree-Grid (Tree-G) were generated by randomly attached cycle motifs and 3-by-3 grids to nodes in 8-level balanced binary trees, respectively.
Since the node features of synthetic graphs had no meaning, the Amazon dataset \cite{rayana2015collective, dou2020enhancing} was used to assess the feature attribution module. Specifically, it consisted of three graphs, wherein links were established based on mutual information between users. The goal was to detect fraudulent users based on given product reviews. Practically, we conducted experiments with all three graphs and selected a graph of users reviewing the same products, which produces the highest recall score, to measure node feature attributions.

\noindent\textbf{Graph Classification.} One synthetic graph and one real-world dataset were used. BA-2motifs (BA-2m) \cite{luo2020parameterized} consists of 1000 graphs with two classes constructed by adding specific motifs to BA graphs, where half contain 5-node house motifs and the other half include 5-node cycle motifs. Next, the Mutag dataset contains 4337 graphs classified into two classes based on their mutagenic effects. It also has ground-truth edge labels pointing out crucial subgraphs linking to mutagenic effects.

\subsection{Baselines}
We compared SCALE with five baseline methods on quantitative and qualitative results to prove the explanation correctness and execution performance. Specifically, baselines are categorized into intrinsically interpretable models and perturbation methods. 

\noindent\textbf{Intrinsically Interpretable Models} use internal model weights to explain predictions directly. We selected four models to compare with SCALE as follows:
\begin{itemize}
    \item \textbf{GCN-MLP:} GCN \cite{kipf2016semi} is not an intrinsically interpretable model. Therefore, we replaced the normalized adjacency matrix with a trainable matrix similar to \cref{graph_classification_mask,node_self_exp} for graph and node classification, respectively. Learnable adjacency matrices were then used to provide structural explanations. 

    \item \textbf{GAT \cite{velickovic2017graph}} can be considered a self-explainable GNN since attention heads capture interactions between nodes. We used three attention heads for each layer and averaged out all heads across layers to provide explanations.

    \item\textbf{SEGNN \cite{dai2021towards}} is a self-explainable GNN based on a similarity module that computes structure distances between an unlabelled node and K-nearest labeled neighbors. We set K such that the out-of-memory problems do not occur and the recall scores are maximized.

    \item\textbf{EGNN \cite{li2022egnn}} is a self-explainable model based on an offline KD paradigm \cite{hinton2015distilling}, which filters out unimportant messages from 2-hop neighbors via two separate masking layers. To construct explanations, we aggregated neighbor scores and selected top K nodes that maximize recall scores. 
\end{itemize}

\noindent\textbf{Perturbation Methods} integrate additional post-hoc training processes specialized for explanation purposes, which follow model-agnostic approaches. We selected two state-of-the-art methods similar to SCALE in problem formulation and explanation model initialization.
\begin{itemize}
    \item \textbf{GNNExplainer \cite{ying2019gnnexplainer}} was the first work that trained edge masks to determine crucial subgraphs based on an information theory approach. However, it has to retrain a mask for each target instance, thus making it less effective for inductive settings and large-scale graphs.
    \item \textbf{PGExplainer \cite{luo2020parameterized}} shared the same approach with GNNExplainer \cite{ying2019gnnexplainer} but initialized masks using embedding vectors from the pre-trained model. Moreover, target instances share trainable weights.
\end{itemize}

\noindent\textbf{Executing \cref{querying_rwr} on GCN-MLP and GAT:} The explanation querying algorithm can operate on various types of pre-trained GNNs as long as these models include normalized adjacency matrices representing interactions between nodes. We named the results of these integration processes SCALE-GCN-MLP and SCALE-GAT, respectively.

\subsection{Evaluation Metrics for Structural Explanations}
Following \cite{ying2019gnnexplainer,luo2020parameterized}, we also formulated explanations as binary classification problems, wherein edges in pre-defined ground-truth motifs were labeled as 1 (positive class), while the other edges belonged to the negative class. Moreover, previous methods used the AUC score as the evaluation metric since learned masks represent probabilities that edges are selected. Instead, we selected precision and recall scores for quantitative comparisons for the following reasons. First, \cref{querying_rwr} cannot be evaluated by the AUC score. Second, we aimed to know the ratio of true positive and false positive edges in explanations varied in particular scenarios. Third, precision and recall scores gave us more information to judge explanation methods. In many cases, explanation methods can achieve high recall scores by including all ground-truth edges but still obtain low precision scores due to numerous false positive edges. Therefore, a good explanation method must provide explanation subgraphs that include all ground-truth edges and contain as few wrong edges as possible, resulting in high scores in both metrics.

\begin{equation}
    \begin{aligned}
        \textrm{Precision} &= \frac{\textrm{True Positive}}{\textrm{True Positive} + \textrm{False Positive}} \\
        \textrm{Recall} &= \frac{\textrm{True Positive}}{\textrm{True Positive} + \textrm{False Negative}}
    \end{aligned}
\end{equation}

\subsection{Configurations}

We followed experimental configurations of \cite{ying2019gnnexplainer, luo2020parameterized}, which used 8:1:1 (train/validation/test) splitting strategy. For fair comparisons, we contacted the authors of GNNExplainer, PGExplainer, and SEGNN to request evaluation scripts for all datasets. Since we did not get responses, we did our best to implement evaluation scripts based on their public source codes. We also followed their published papers and source codes to train baselines on datasets. Moreover, we used Youden's J Statistic for determining selection thresholds in baselines that output edge selection probabilities. According to \cite{ying2019gnnexplainer, luo2020parameterized}, explained instances were manually selected, regardless of which set they belonged to in training. Please refer to our GitHub code at \href{https://github.com/alexbui91/SCALE}{https://bit.ly/SCALEGNN} for more detail.

\begin{table}[h]
    \centering
    \begin{tabular}{c|c|c|c|c|c}
    \hline
      & \thead{MLP\\Layers} & \thead{GCN\\Layers} & \thead{Hidden\\Size} & $\lambda$ & \thead{Num.\\ Epochs}\\
      \hline
    Amazon & 2 & 2 & 32 & 0.1 & 200\\
    BA-Shapes & 3 & 6 & 32 & 0.1 & 1000\\ 
    BA-Community & 3 & 6 & 64 & 0.1 & 1000 \\ 
    Tree-Cycle & 3 & 6 & 64 & 0.1 & 1000\\
    Tree-Grid & 3 & 6 & 64 & 0.1 & 1000\\
    \hline
    BA-2motifs & 3 & 4 & 64 & 4 & 200 \\
    Mutag & 3 & 4 & 64 & 4 & 200 \\
    \hline
    \end{tabular}
    \vspace{0.1cm}
    \caption{Training Hyper-parameters}
    \label{tab:hyper}
\end{table}

SCALE modules were trained with hyper-parameter settings presented in \cref{tab:hyper}. The hidden size represents the dimension of transformation matrices in GNN-based models and denotes the first layer's dimension in MLP models. The sizes of the last layers in MLP models depend on the model roles, which can be 1 in mask initialization or 2 in classification tasks. Furthermore, the hidden size was reduced by half after each layer. For instance, $[64,32,2]$ denotes an MLP with three layers, with the first layer containing 64 hidden units.
Practically, we used the learning rate of 0.01 and set $\tau$ as 2 in all settings. The jumping probability $d$ was set as 0.9 for Tree-Grid and 0.55 for others. In training, the magnitude of $\lambda$ correlated with the amount of knowledge distilled from the teacher to the student. We studied the impacts of $\lambda$ and $d$ on explanation correctness in ablation studies.

We ran SCALE and baselines five times in each dataset using a machine with one NVIDIA Tesla V100 16GB GPU and reported the average results. We used PyTorch v1.10.2 and DGL v0.9.0 for building models in SCALE. Other baselines except PGExplainer were also executed using the same PyTorch version. We used Tensorflow v2.9.1 for experiments with PGExplainer. DeepLIFT was executed using a PyTorch API provided by \cite{shapDeep19:online}.

\begin{table*}[hbt]
    \centering
    \begin{tabular}{c|c|c|c|c|c|c|c|c|c|c|c|c}
    \hline
    & \multicolumn{2}{c|}{BA-Shapes} & \multicolumn{2}{c|}{BA-Community} & \multicolumn{2}{c|}{Tree-Cycle} & \multicolumn{2}{c|}{Tree-Grid} & \multicolumn{2}{c|}{BA-2motifs} & \multicolumn{2}{c}{Mutag}\\
    \cline{2-13}
    & P & R & P & R & P & R & P & R & P & R & P & R \\
    \hline
    GCN-MLP & 94.03 & 52.50 & 65.51 & 92.08 & 68.75 & 85.56 & 48.73 & 69.13 & 21.66 & \textbf{100} & 13.71 & 66.67 \\
    GAT & 89.35 & 99.03 & 72.19 & 97.36 & 68.16 & 75.83 & 54.71 & 30.14 & - & - & - & - \\
    SEGNN & 98.55 & 49.38 & \underline{97.39} & 46.46 & 69.79 & 82.78 & 76.35 & 73.33
     & - & - & - & - \\
    EGNN & 56.55 & \underline{99.44} & 42.73 & 93.89 & 62.17 & 66.67 & 77.85 & 63.03
     & - & - & - & - \\
    \hline
    GNNExplainer & 80.44 & \textbf{100} & 59.27 & \textbf{100} & 73.92 & 87.34 & 75.18 & 53.79 & 26.46 & \underline{94.18} & 14.66 & 71.49 \\
    PGExplainer & 96.85 & \textbf{100} & 54.97 & 97.07 & 99.25 & \underline{99.57} & 91.05 & \underline{87.81} & \underline{95.58} & \textbf{100} & \underline{50.64} & \underline{99.37} \\
    \hline

    SCALE & \textbf{98.90} & \textbf{100} & \textbf{99.17} & \textbf{100} & \textbf{99.45} & \textbf{100} & \textbf{97.11} & \textbf{91.00} & \textbf{96.25} & \textbf{100} & \textbf{66.18} & \textbf{99.72} \\ 
    SCALE-GCN-MLP & \underline{98.87} & \textbf{100} & 85.75 & \underline{98.61} & \underline{99.36} & \textbf{100} & \underline{94.90} & 84.31 & - & - & - & - \\
    SCALE-GAT & 98.63 & \textbf{100} & 91.89 & 94.44 & 83.59 & 79.02 & 84.11 & 64.84 & - & - & - & - \\
    \hline
    \end{tabular}
    \vspace{0.1cm}
    \caption{Quantitative Comparison of SCALE and baselines on structural explanation correctness. SCALE is superior to baselines, especially intrinsically interpretable models. Explanation results are significantly improved when executing \cref{querying_rwr} on GCN-MLP and GAT. Here, P is short for the precision score, and R denotes the recall score.}
    \label{tab:quant_compare}
\end{table*}

\section{Experimental Results}\label{exp_results}

\subsection{Quantitative Comparison with Baselines}

First, we compared methods on the correctness of structural explanations. From \cref{tab:quant_compare}, we have the following observations. SCALE is superior to all baselines in explaining both node and graph predictions. Specifically, it achieves outstanding precision and recall scores in node classification datasets and outperforms state-of-the-art methods GNNExplainer and PGExplainer. GCN-MLP, GAT, and EGNN achieve high recall scores on BA-based datasets since ground-truth motifs only need at most 2-hop traversals. However, SEGNN performs poorly on these datasets since the number of sampling hops cannot be larger than one due to out-of-memory errors. In tree-based datasets, self-explainable models are inferior to SCALE due to the ineffectiveness of their explanation procedures. Perturbation-based methods follow sampling-then-choosing approaches, which can cause multiple false edges to be included in explanations. Conversely, SCALE achieves high precision scores since it expands explanation motifs from target nodes until meeting vertex thresholds corresponding to ground-truth motifs based on \cref{querying_rwr}. SCALE outperforms baselines on Mutag, wherein the precision score gains are 15.54\% compared to PGExplainer and 51.52\% in comparison with GNNExplainer. Its performance is also comparable to PGExplainer on the BA-2motifs dataset. 

\begin{table}[ht]
    \centering
    \begin{tabular}{c|c|c|c|c|c|c}
        \hline
         & BA-S & BA-C & Tree-C & Tree-G & BA-2m & Mutag \\
         \hline
         GCN-MLP & 0.16 & 0.22 & 0.20 & 0.89 & 0.42 & 2.49 \\
         GAT & 0.16 & 0.20 & 0.18 & 1.11 & - & -\\
         SEGNN & 0.24 & 0.26 & 0.33 & 1.59 & - & -\\
         EGNN & 13.52 & 19.60 & 15.43 & 23.05 & - & -\\
         \hline
         GNNExpl. & 40.79 & \underline{40.77} & \underline{34.11} & \underline{155.35} & \underline{107.42} & 630.42 \\
         PGExpl. & \underline{29.33} & 167.89 & 55.61 & 515.16 & 183.4 & \underline{153.2}\\
         \hline
         SCALE & \textbf{1.58} & \textbf{1.62} & \textbf{2.17} & \textbf{5.81} & \textbf{1.53} & \textbf{6.70} \\
         \hline
    \end{tabular}
    \caption{Comparing Execution Time on Explanation Set Between SCALE and Baselines. Results are measured in seconds. }
    \label{tab:time_compare}
\end{table}

Secondly, we aim to demonstrate the superiority of SCALE in running performance. \cref{tab:time_compare} shows that SCALE is significantly faster than post-hoc explanation methods in all experiments. The performance gains are up to 94x compared to GNNExplainer and 120x in comparison with PGExplainer. Even though SCALE is slightly slower than self-explainable baselines in some scenarios, the gaps are insignificant. Moreover, this drawback is acceptable, considering SCALE's outstanding explanation scores compared with these methods.

\subsection{Qualitative Comparison with Baselines}

\begin{figure*}[ht]
    \centering
    \setlength\tabcolsep{3pt}
    \renewcommand{\arraystretch}{1}
    \begin{tabular}{r p{2.0cm}p{2.0cm}p{2.0cm}p{2.0cm}p{2.0cm}p{2.0cm}}
         \textbf{Ground Truth} & 
         \multicolumn{1}{m{2.0cm}}{\includegraphics[clip,trim=2.5cm 2.5cm 2.0cm 2.4cm,width=2.0cm]{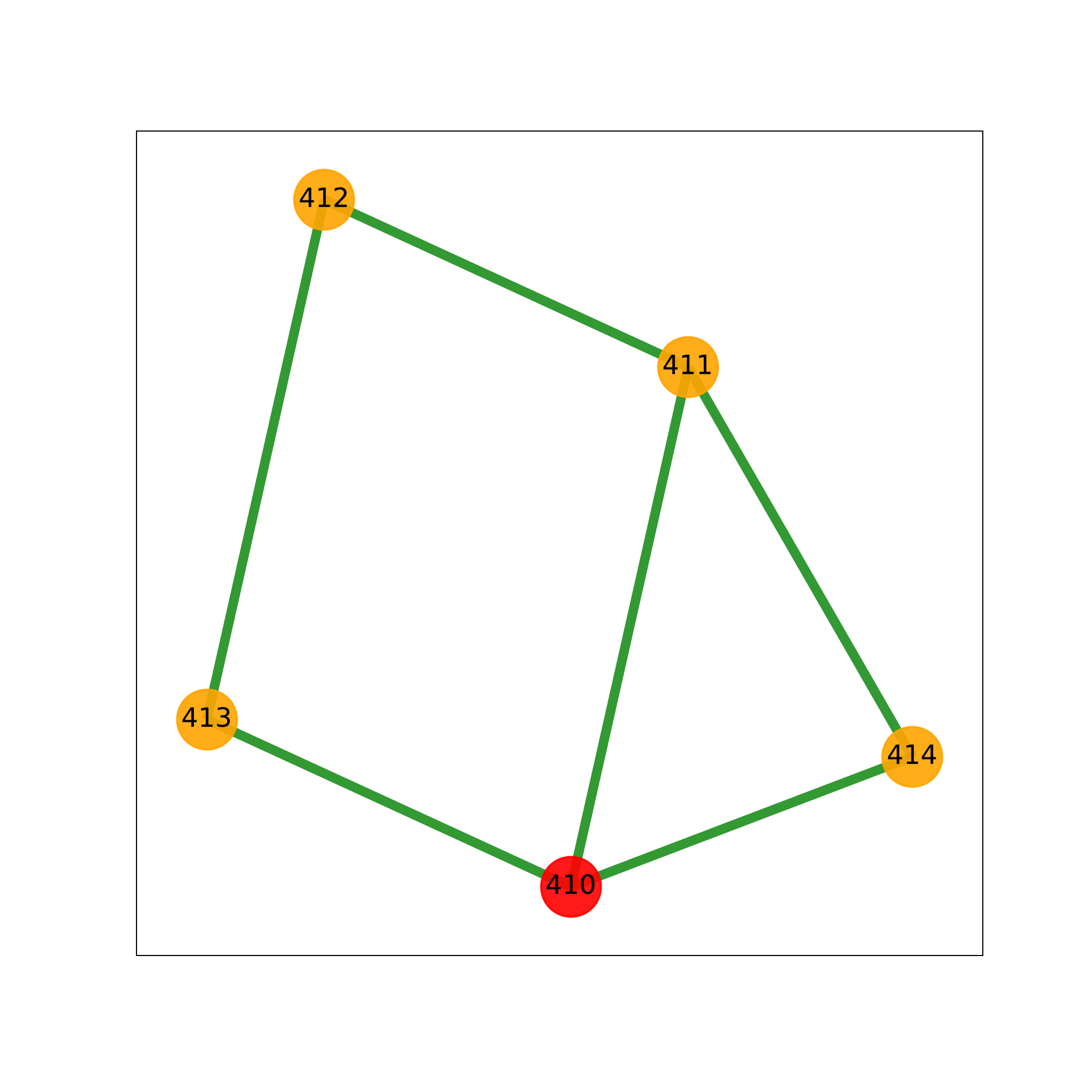}} & 
        \multicolumn{1}{m{2.0cm}}{\includegraphics[clip,trim=2.5cm 2.5cm 2.0cm 2.4cm,width=2.0cm]{figures/graphs/ba_shape_410_real.pdf}} & 
        \multicolumn{1}{m{2.0cm}}{\includegraphics[clip,trim=2.5cm 2.5cm 2.0cm 2.4cm,width=2.0cm]{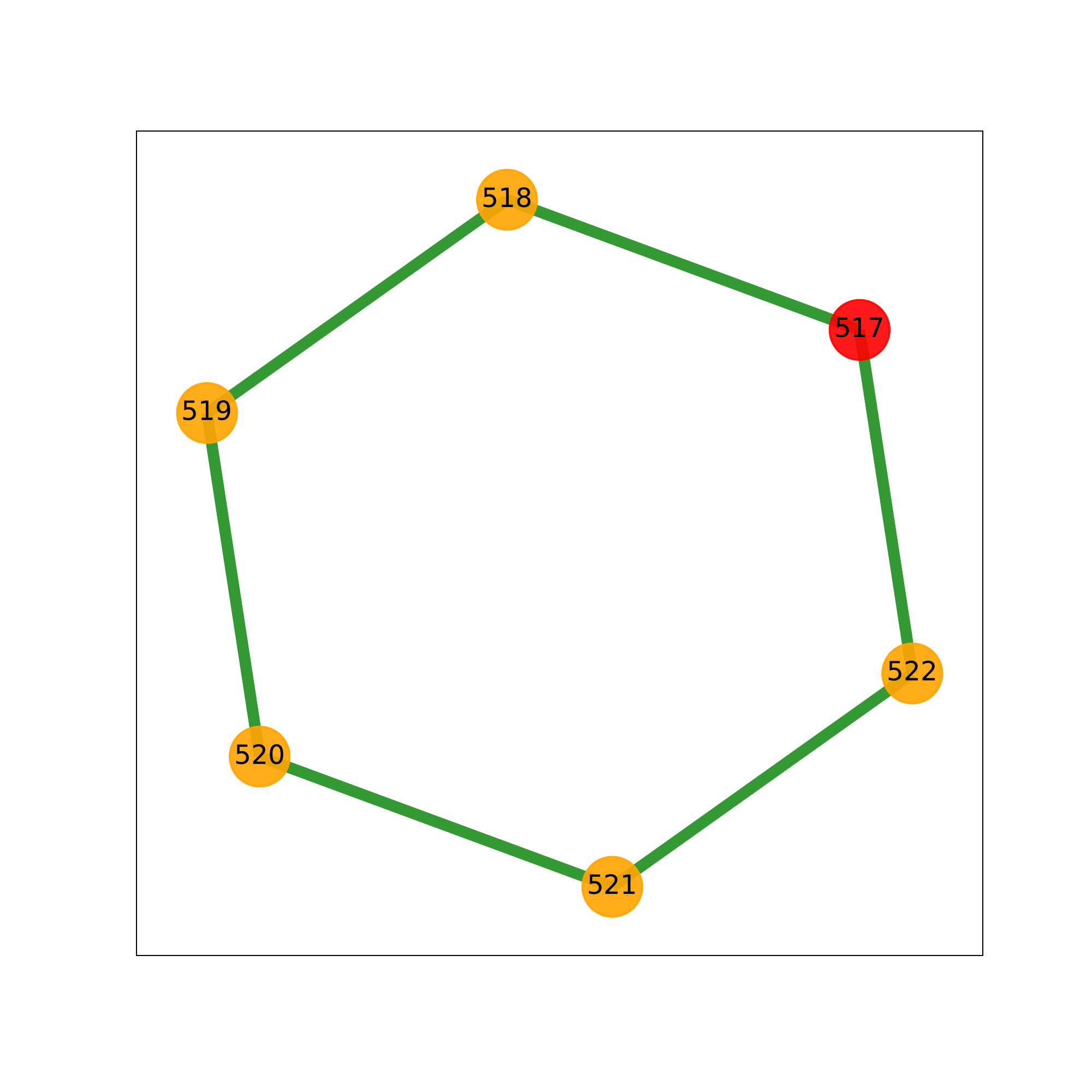}} & 
        \multicolumn{1}{m{2.0cm}}{\includegraphics[clip,trim=2.5cm 2.5cm 2.0cm 2.4cm,width=2.0cm]{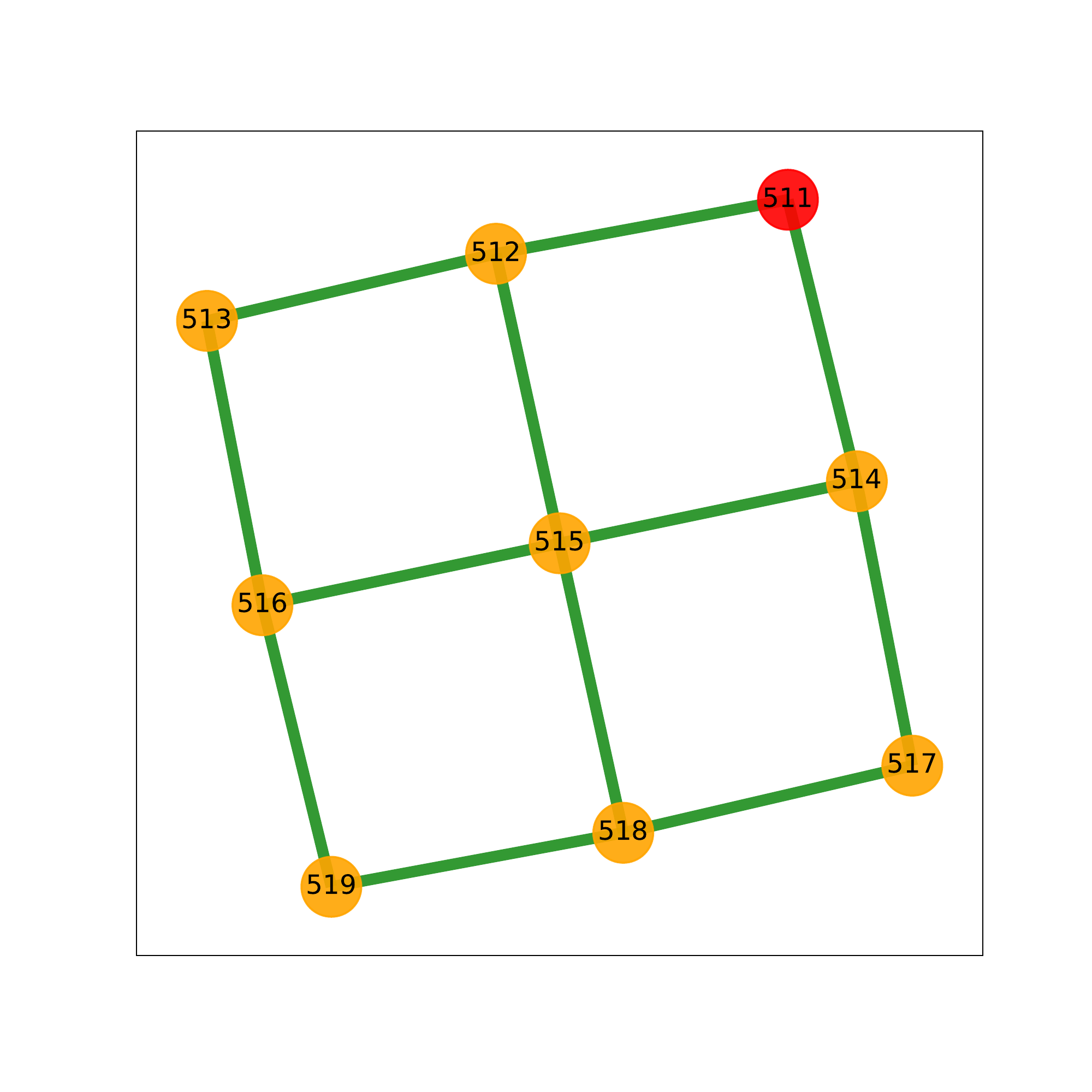}} & 
        \multicolumn{1}{m{2.0cm}}{\includegraphics[clip,trim=2.5cm 2.5cm 2.0cm 2.4cm,width=2.0cm]{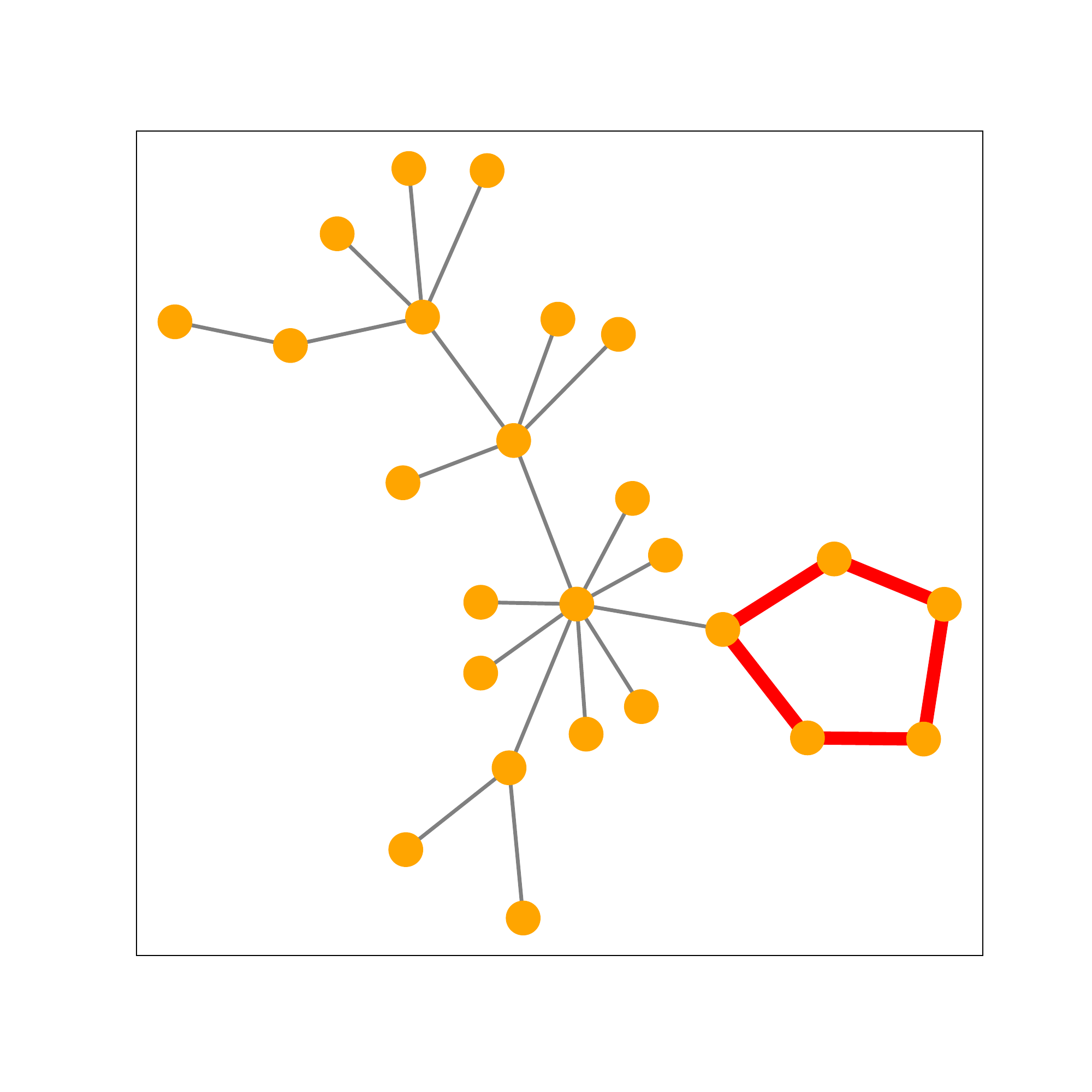}} & 
        \multicolumn{1}{m{2.0cm}}{\includegraphics[clip,trim=2.5cm 2.5cm 2.0cm 2.4cm,width=2.0cm]{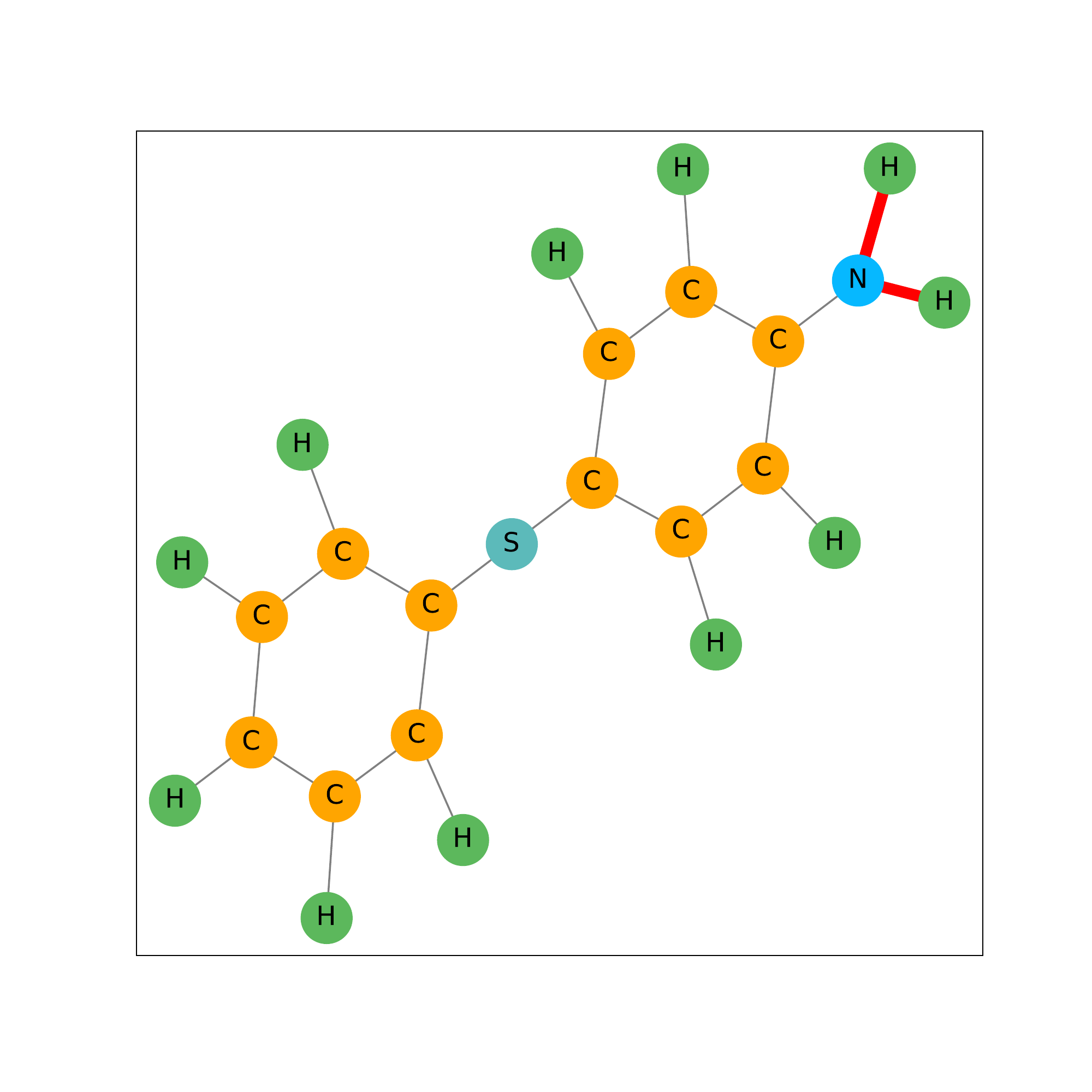}}
        \\
         \textbf{GNNExplainer} & 
         \multicolumn{1}{m{2.0cm}}{\includegraphics[clip,trim=2.5cm 2.5cm 2.0cm 2.4cm,width=2.0cm]{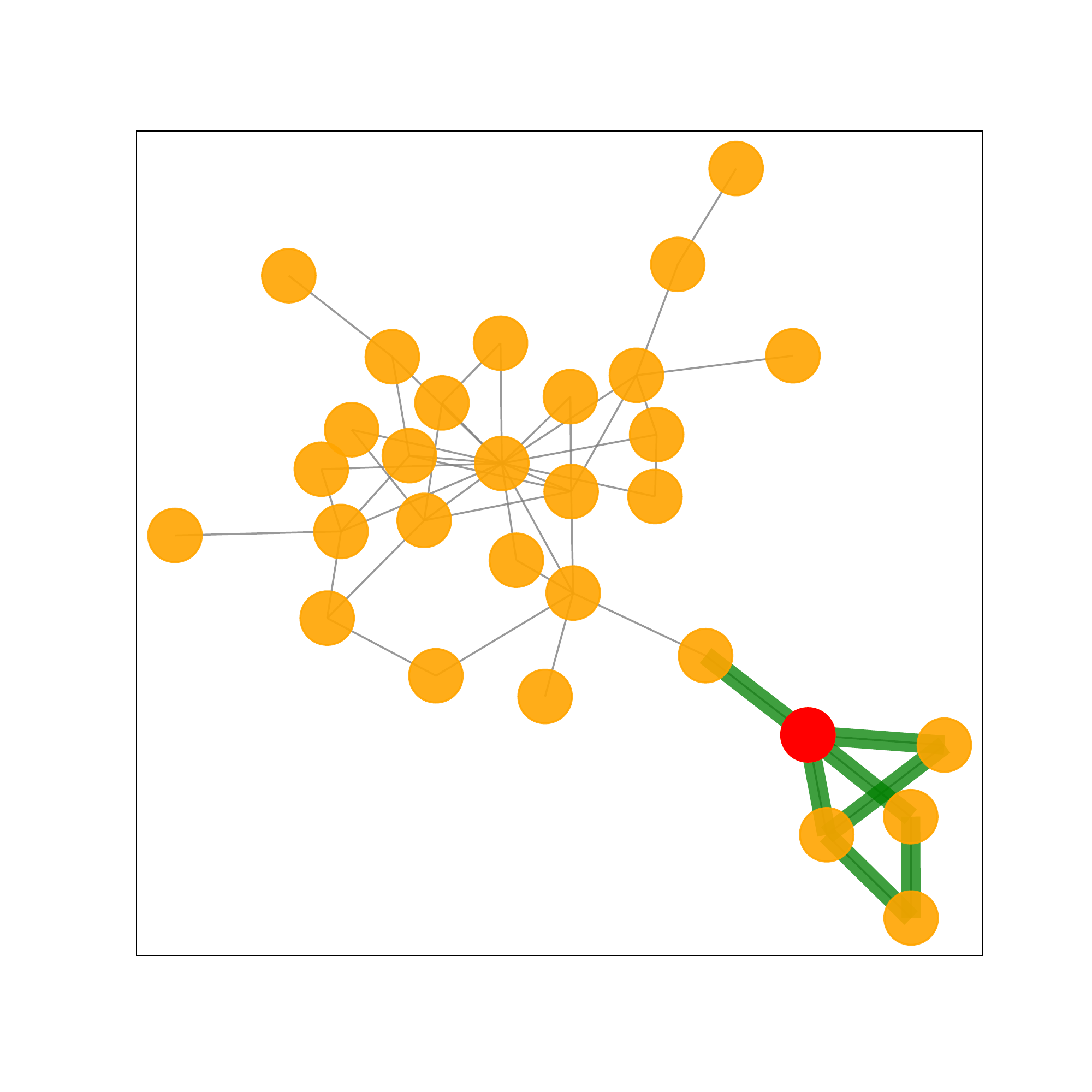}} & 
        \multicolumn{1}{m{2.0cm}}{\includegraphics[clip,trim=2.5cm 2.5cm 2.0cm 2.4cm,width=2.0cm]{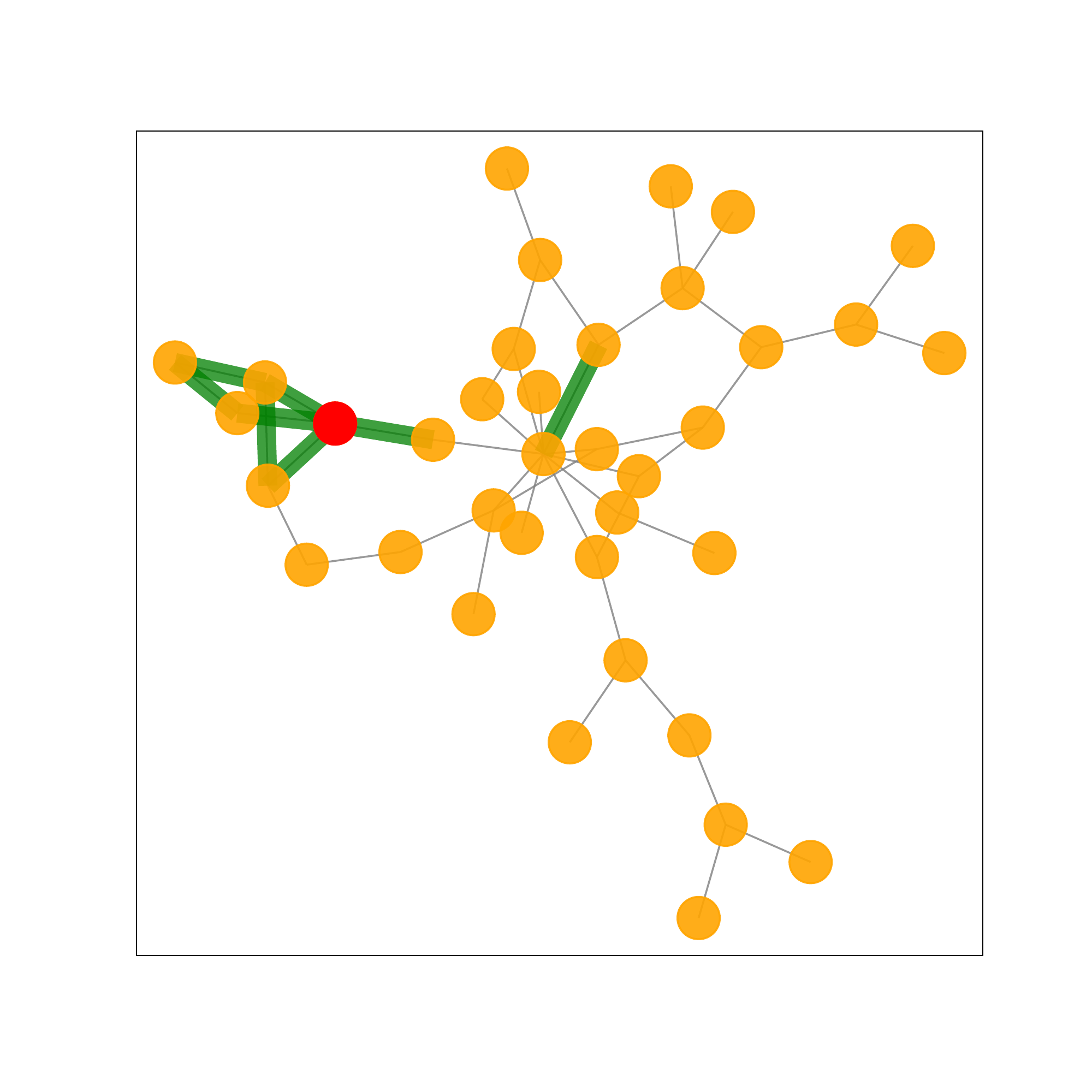}} & 
        \multicolumn{1}{m{2.0cm}}{\includegraphics[clip,trim=2.5cm 2.5cm 2.0cm 2.4cm,width=2.0cm]{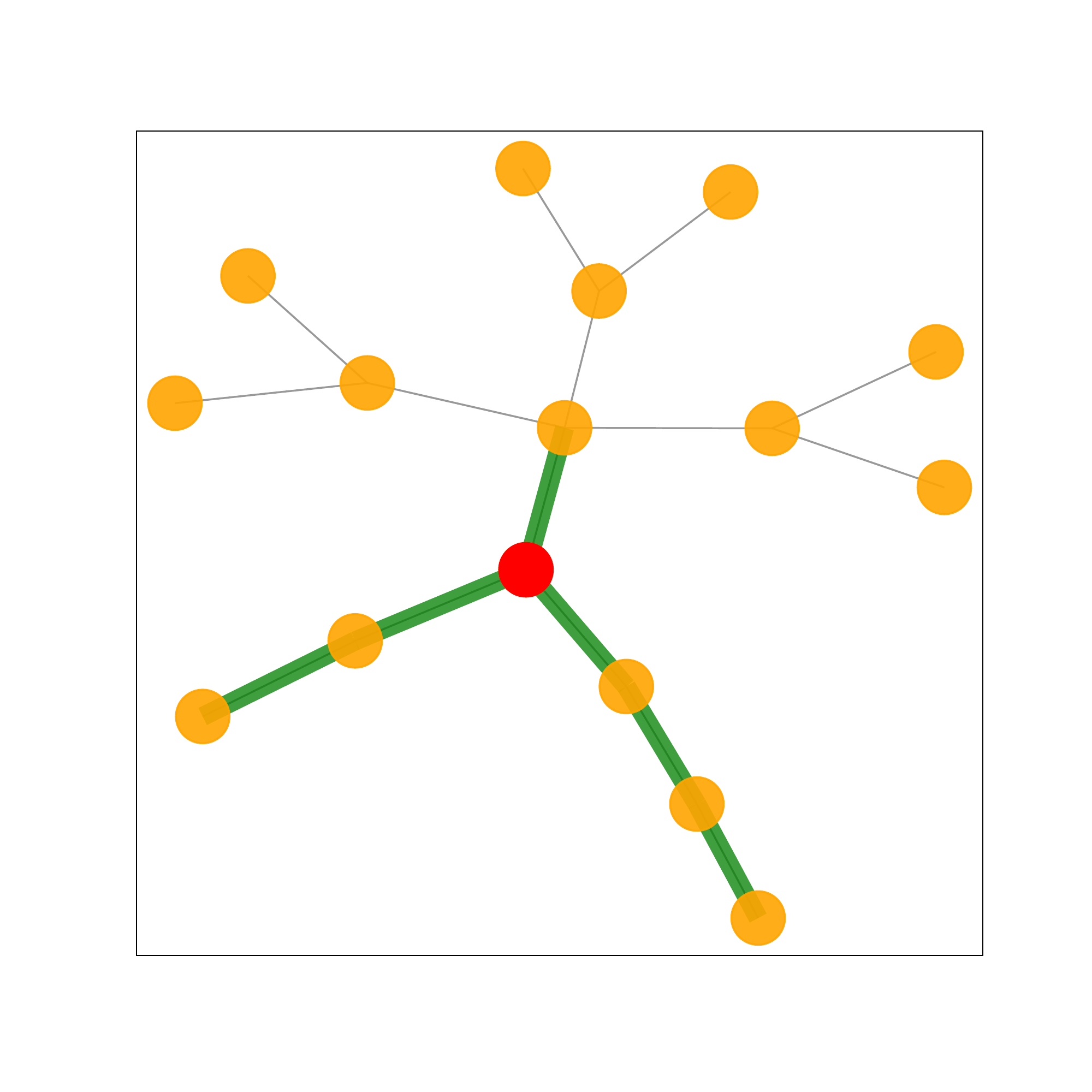}} & 
        \multicolumn{1}{m{2.0cm}}{\includegraphics[clip,trim=2.5cm 2.5cm 2.0cm 2.4cm,width=2.0cm]{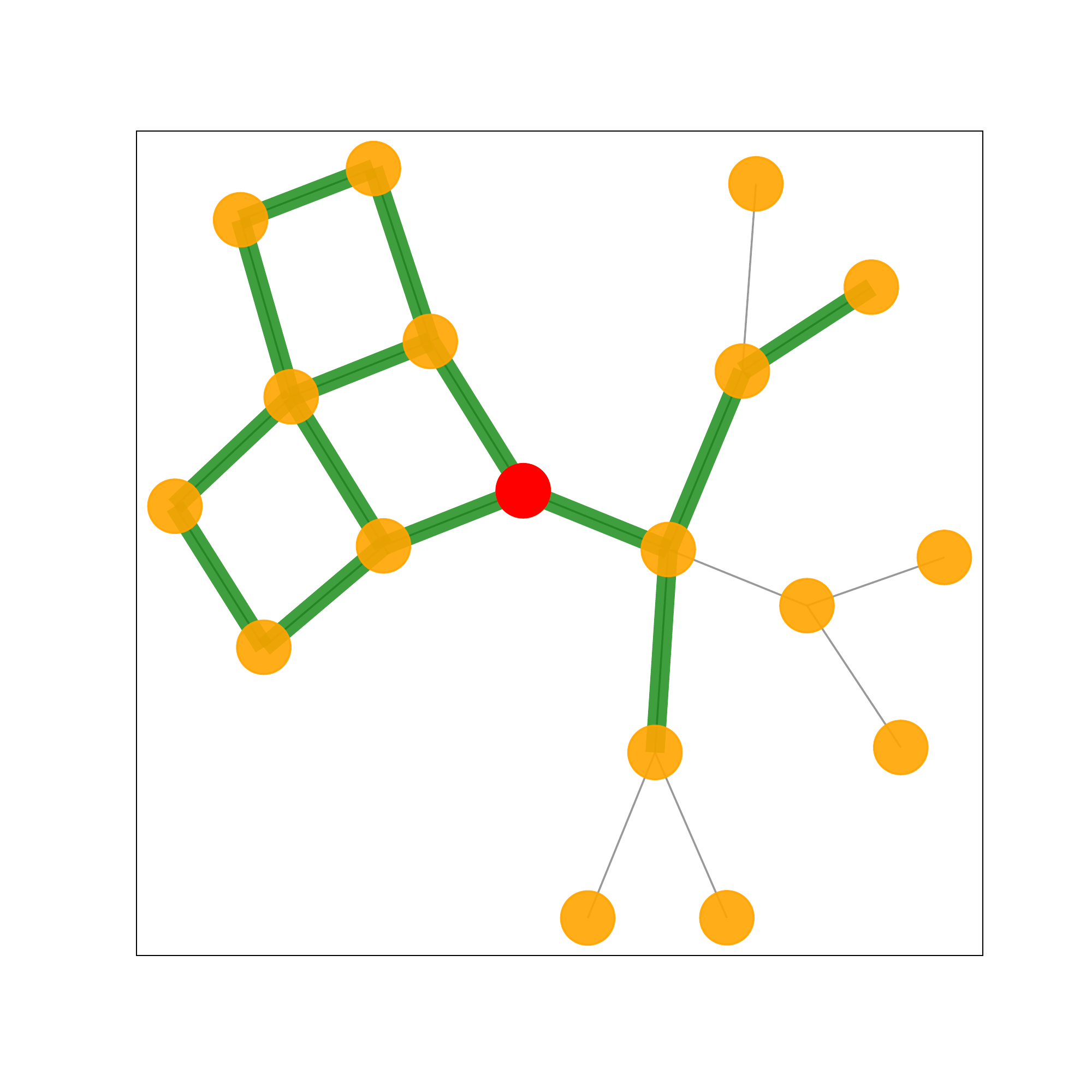}} & 
        \multicolumn{1}{m{2.0cm}}{\includegraphics[clip,trim=2.5cm 2.5cm 2.0cm 2.4cm,width=2.0cm]{figures/graphs/pg_ba2motif_5.pdf}} & 
        \multicolumn{1}{m{2.0cm}}{\includegraphics[clip,trim=2.5cm 2.5cm 2.0cm 2.4cm,width=2.0cm]{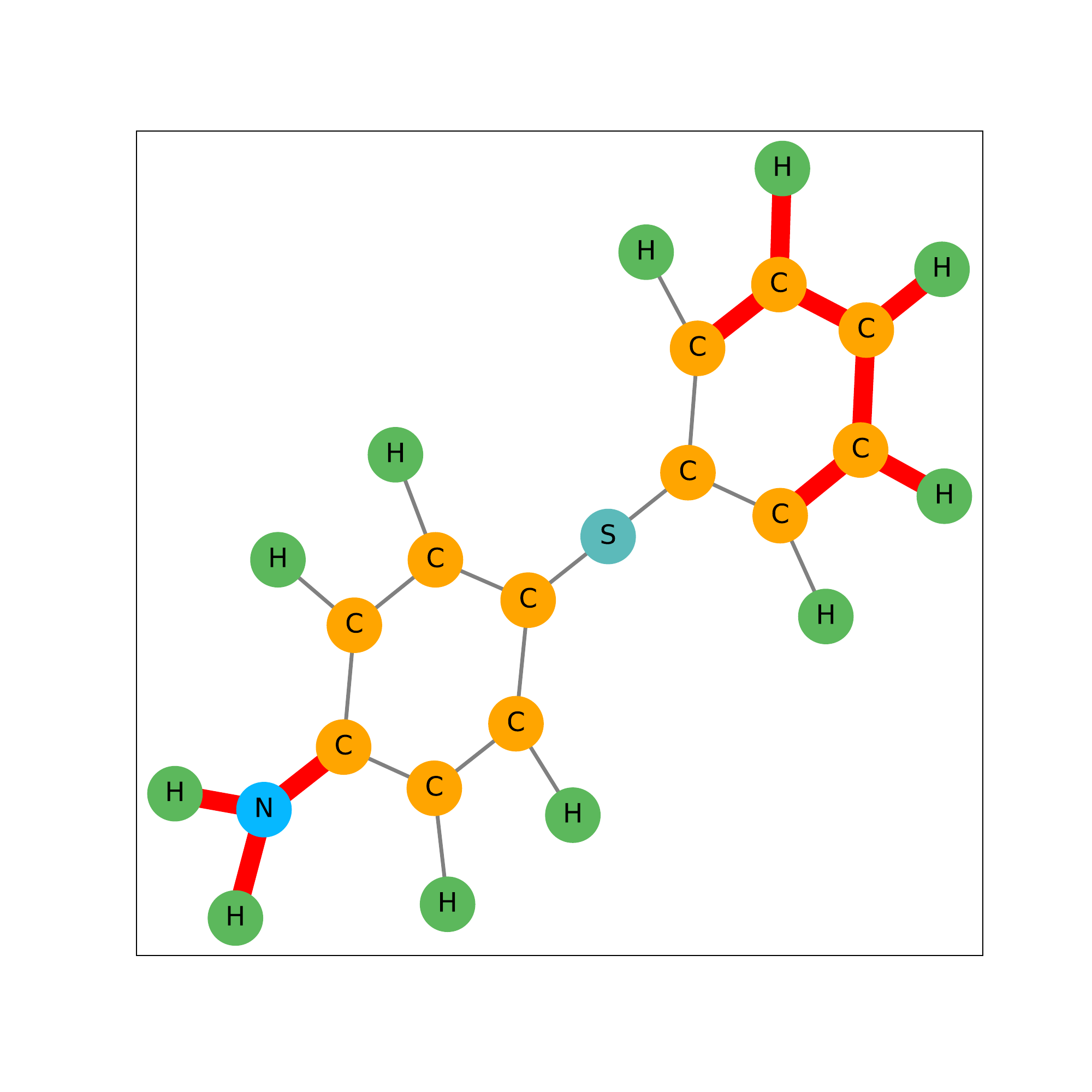}}
         \\
        
        \textbf{PGExplainer} & 
         \multicolumn{1}{m{2.0cm}}{\includegraphics[clip,trim=2.5cm 2.5cm 2.0cm 2.4cm,width=2.0cm]{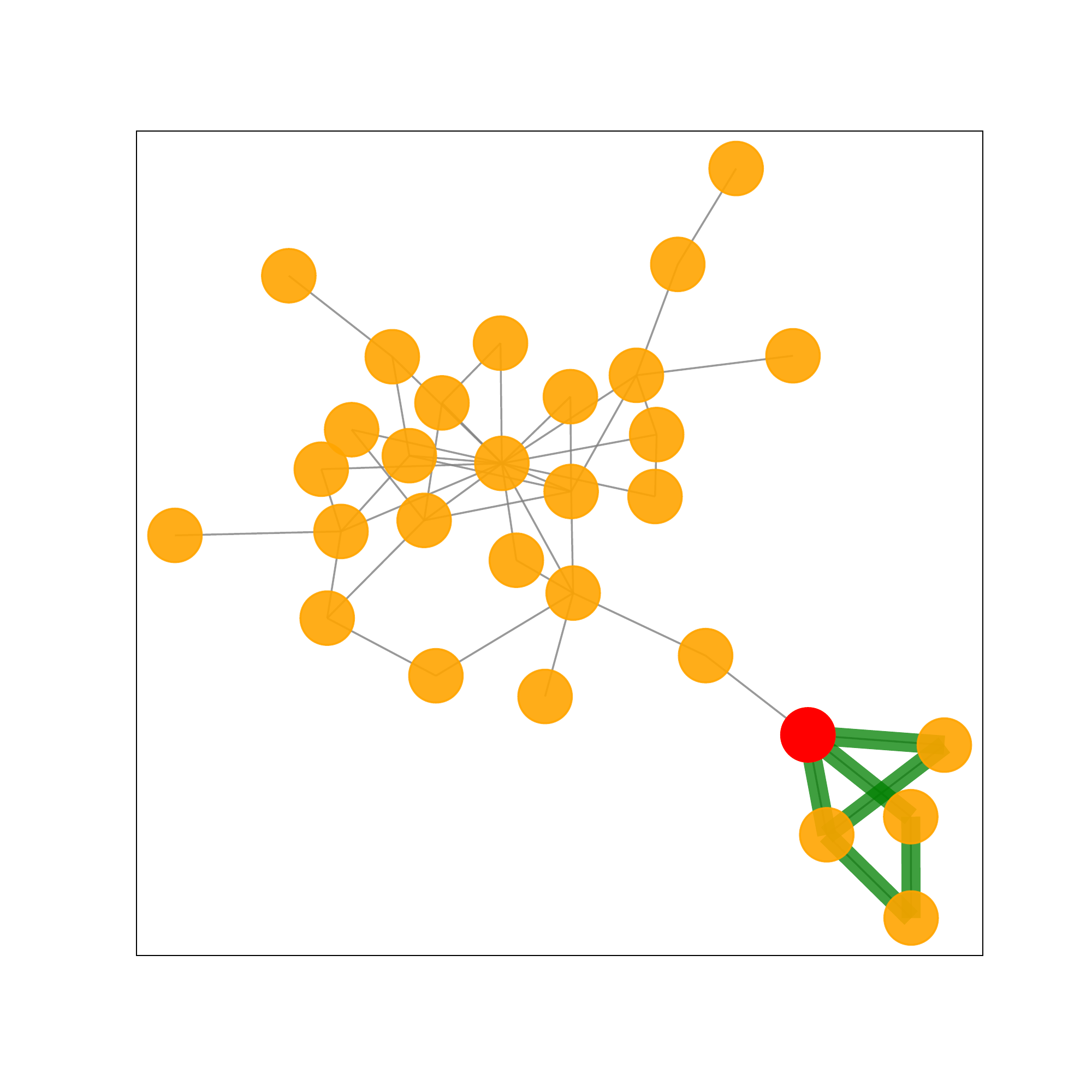} } & 
        \multicolumn{1}{m{2.0cm}}{\includegraphics[clip,trim=2.5cm 2.5cm 2.0cm 2.4cm,width=2.0cm]{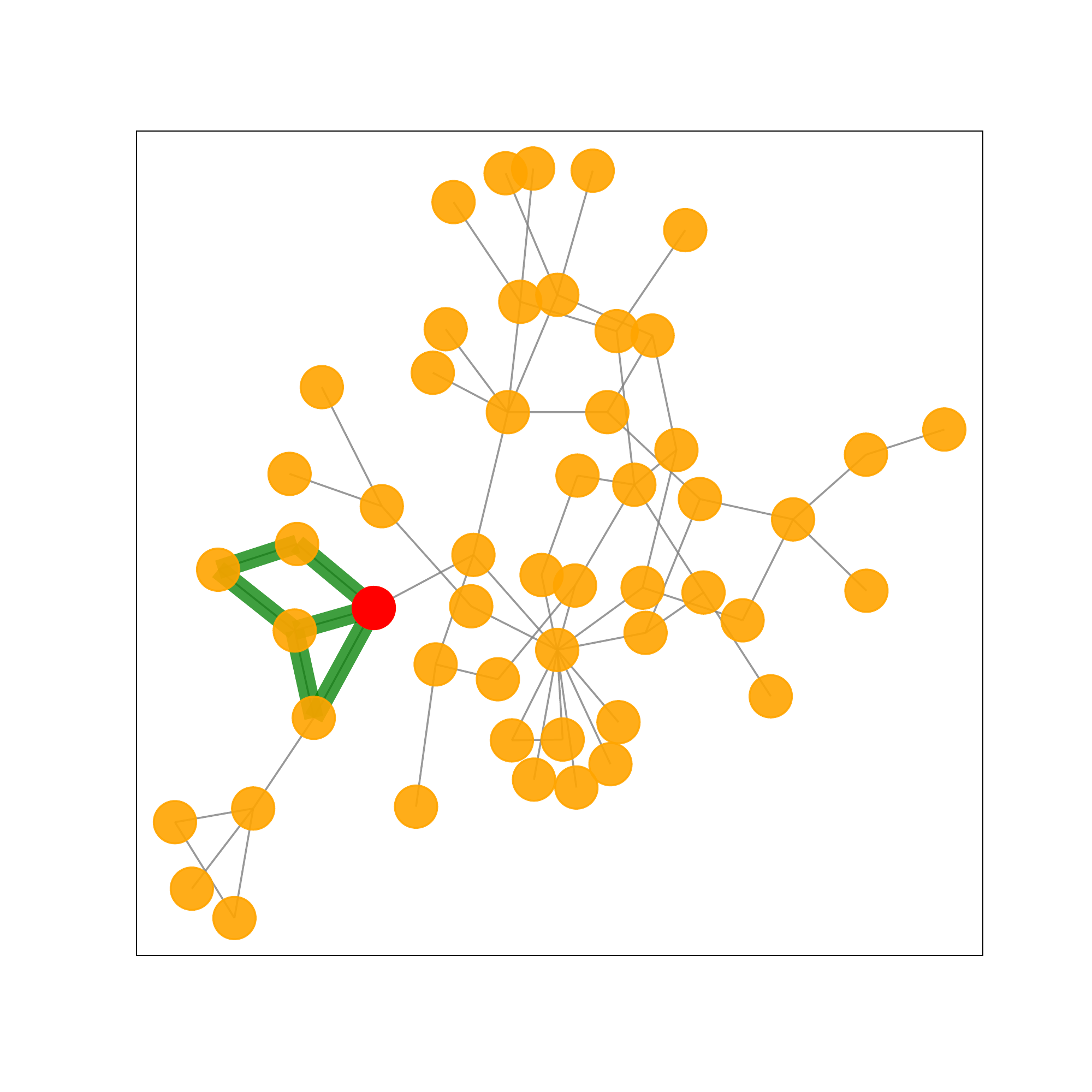} } & 
        \multicolumn{1}{m{2.0cm}}{\includegraphics[clip,trim=2.5cm 2.5cm 2.0cm 2.4cm,width=2.0cm]{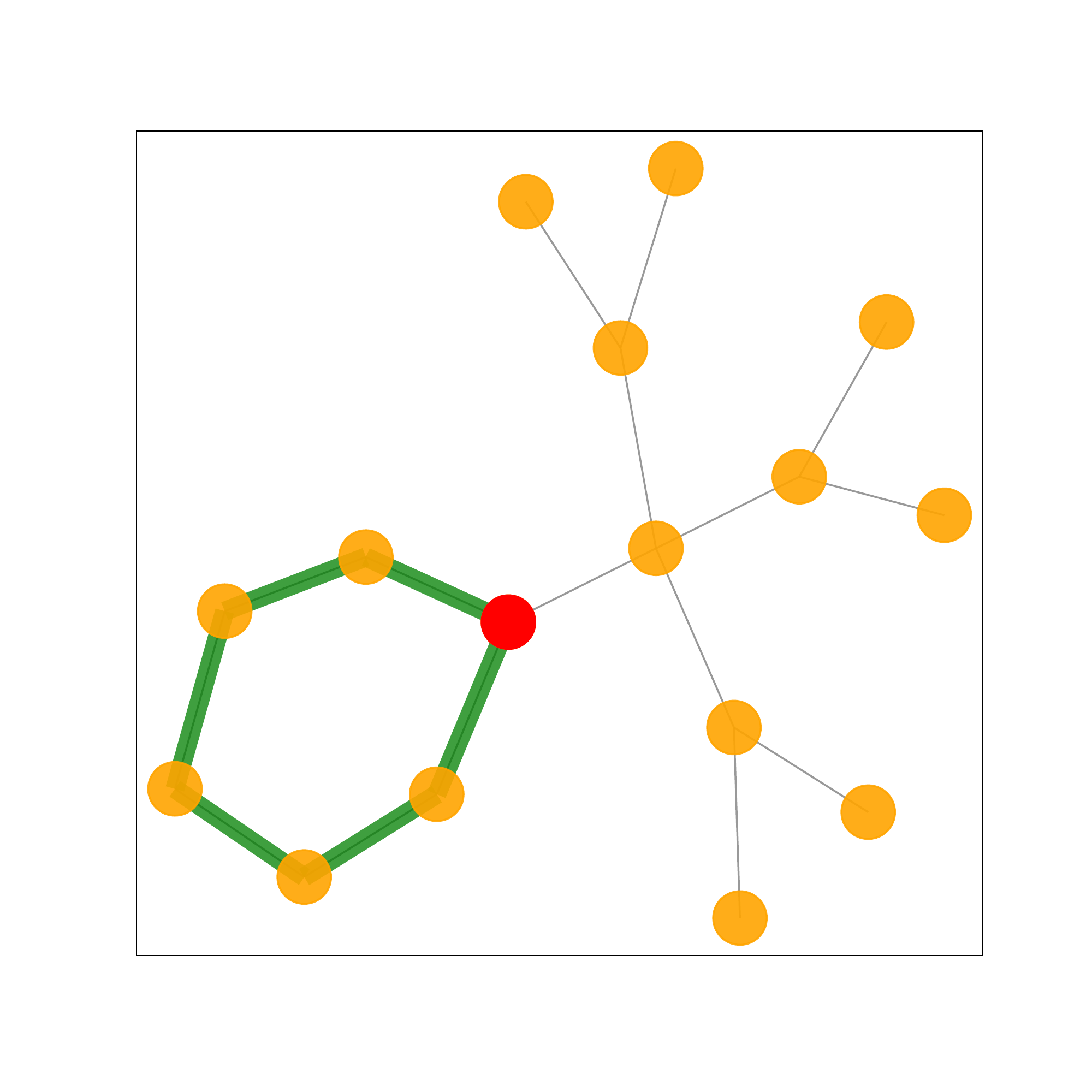} } & 
        \multicolumn{1}{m{2.0cm}}{\includegraphics[clip,trim=2.5cm 2.5cm 2.0cm 2.4cm,width=2.0cm]{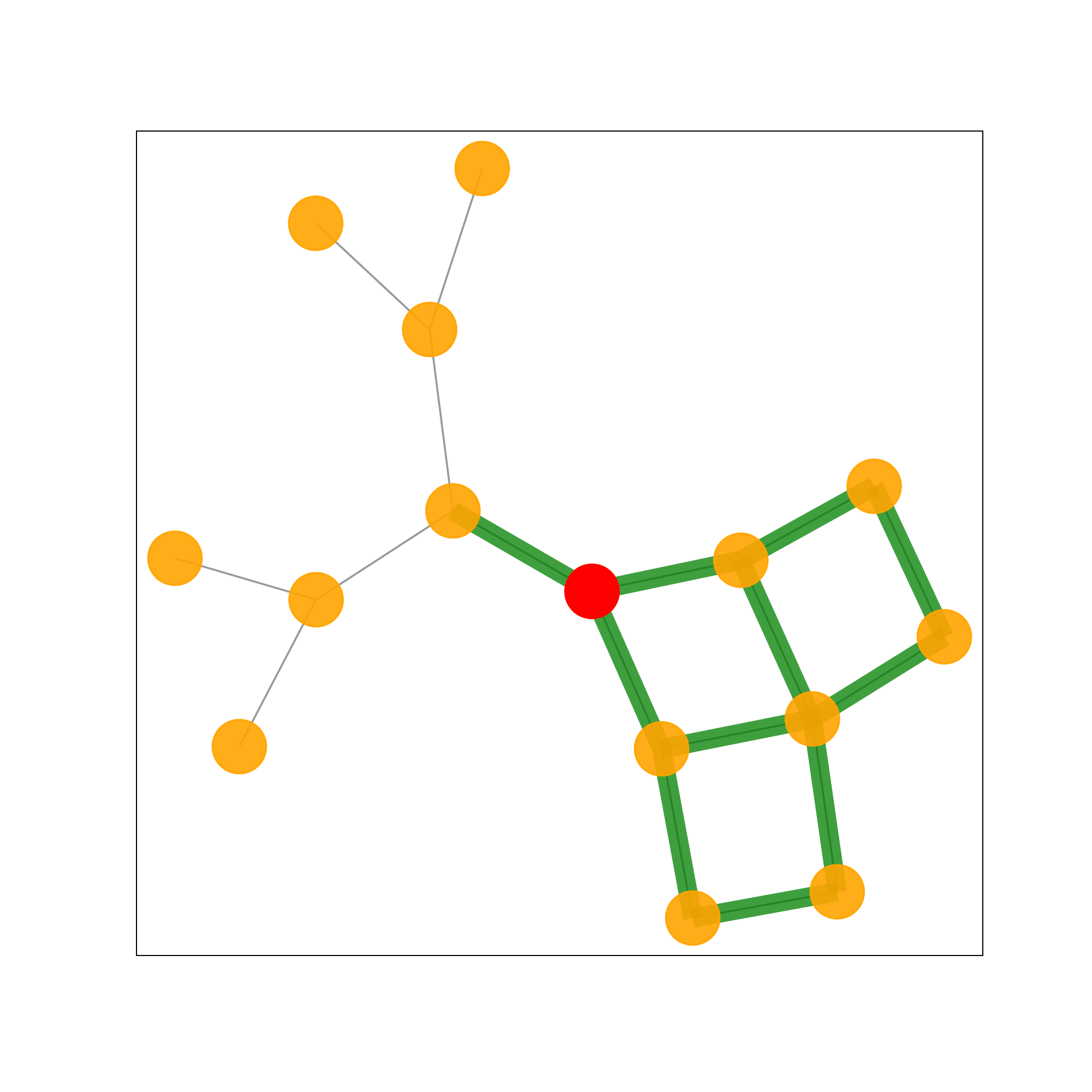} } & 
        \multicolumn{1}{m{2.0cm}}{\includegraphics[clip,trim=2.5cm 2.5cm 2.0cm 2.4cm,width=2.0cm]{figures/graphs/pg_ba2motif_5.pdf} } & 
        \multicolumn{1}{m{2.0cm}}{\includegraphics[clip,trim=2.5cm 2.5cm 2.0cm 2.4cm,width=2.0cm]{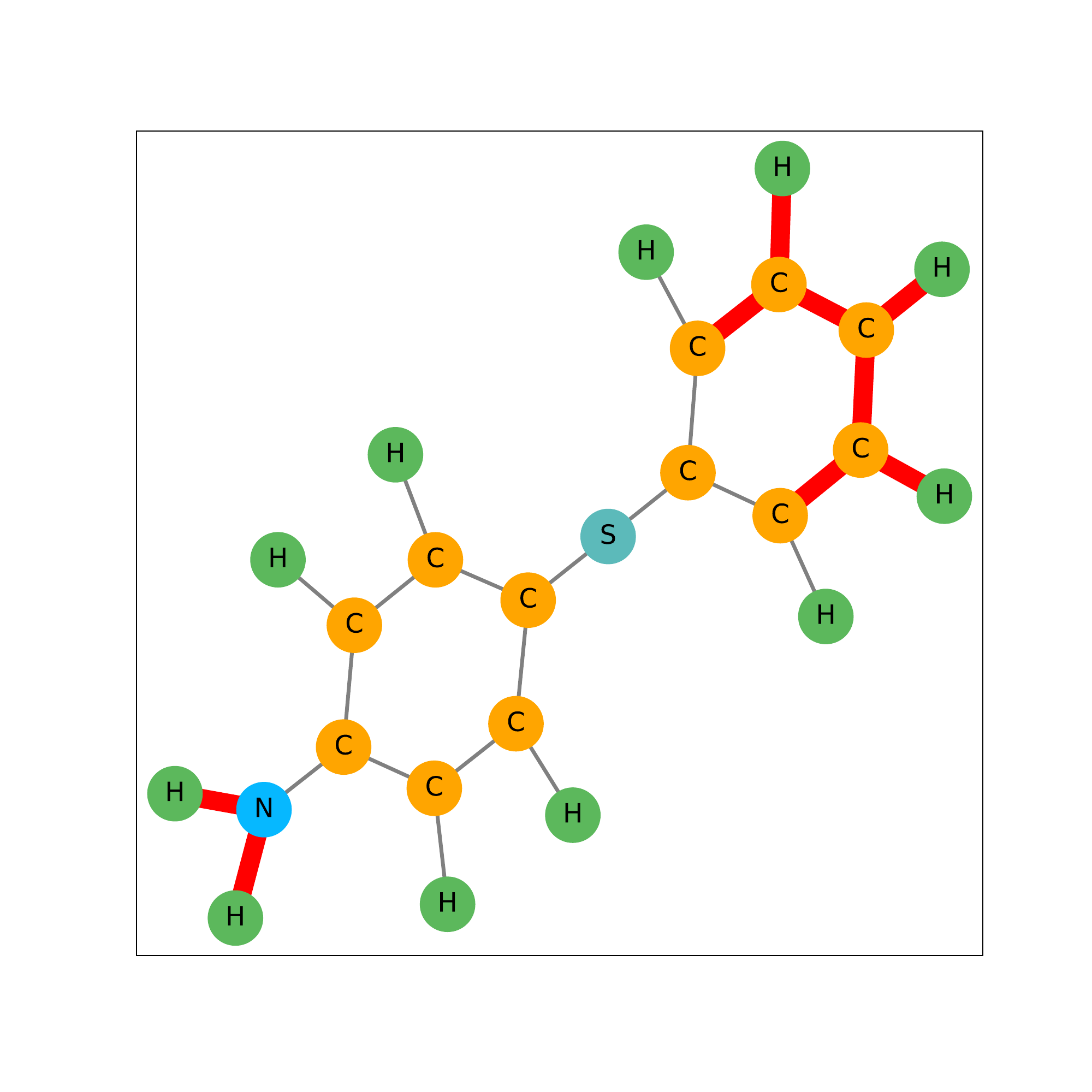}}
         \\

         \textbf{SCALE} & 
         \multicolumn{1}{m{2.0cm}}{\includegraphics[clip,trim=2.5cm 2.5cm 2.0cm 2.4cm,width=2.0cm]{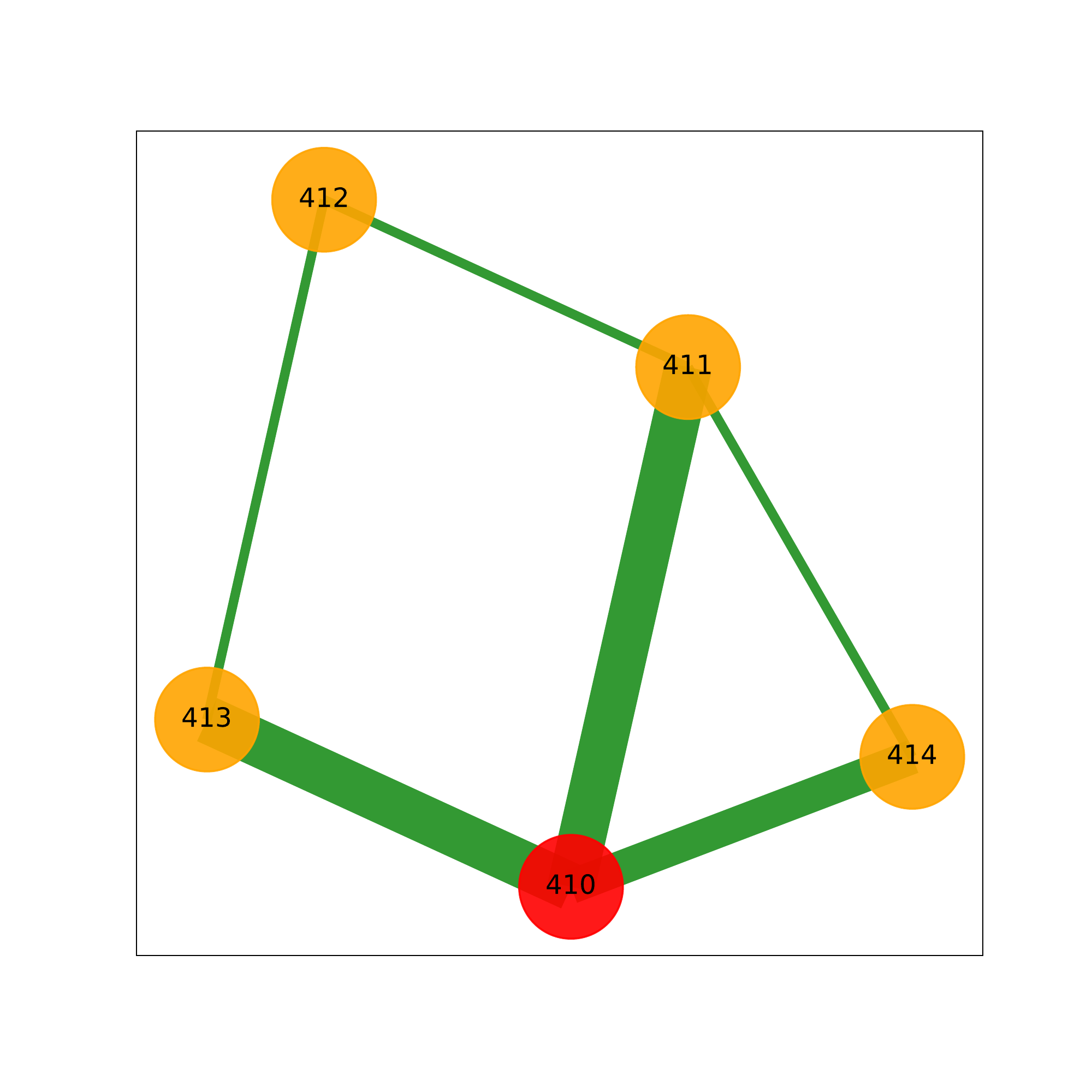}} & 
        \multicolumn{1}{m{2.0cm}}{\includegraphics[clip,trim=2.5cm 2.5cm 2.0cm 2.4cm,width=2.0cm]{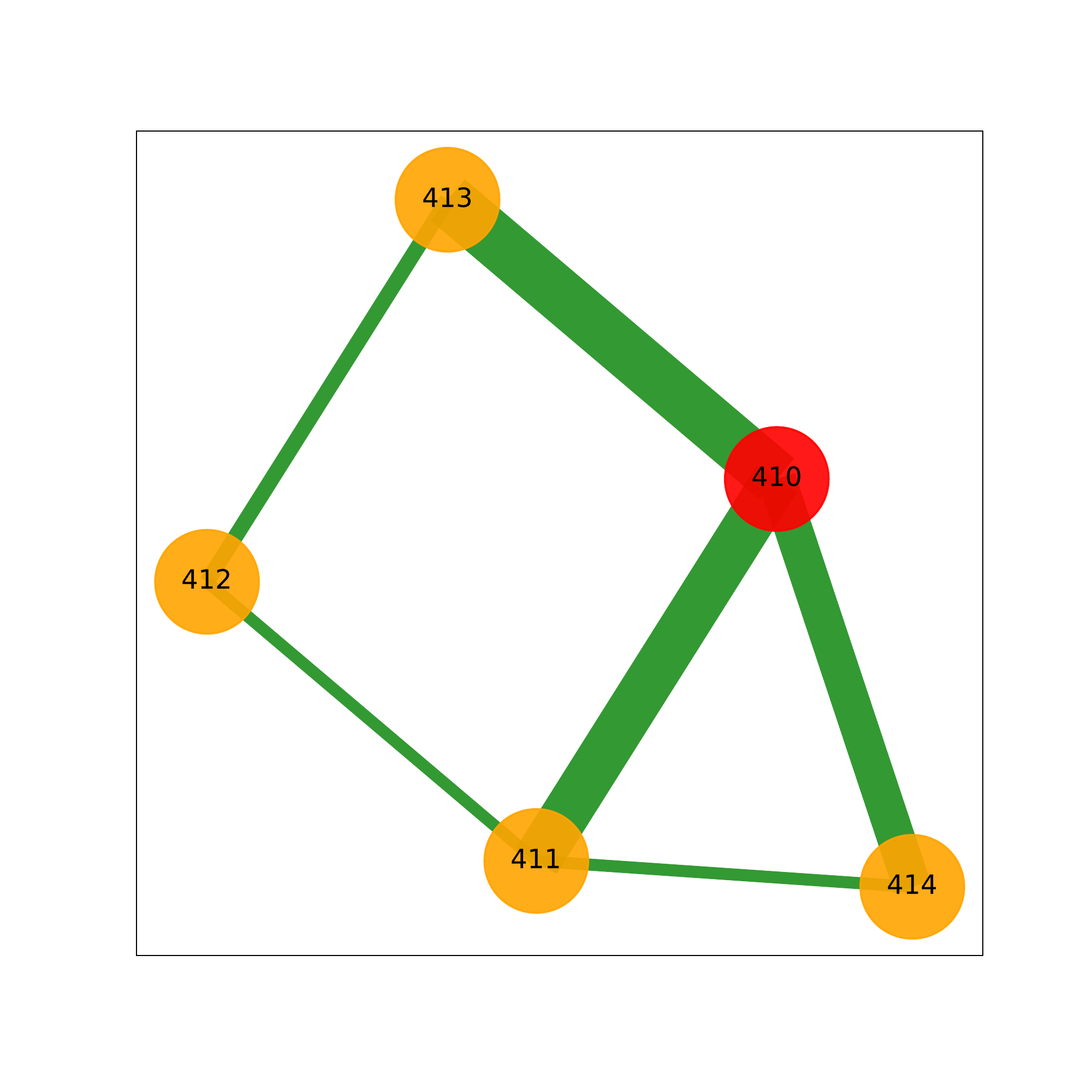}} & 
        \multicolumn{1}{m{2.0cm}}{\includegraphics[clip,trim=2.5cm 2.5cm 2.0cm 2.4cm,width=2.0cm]{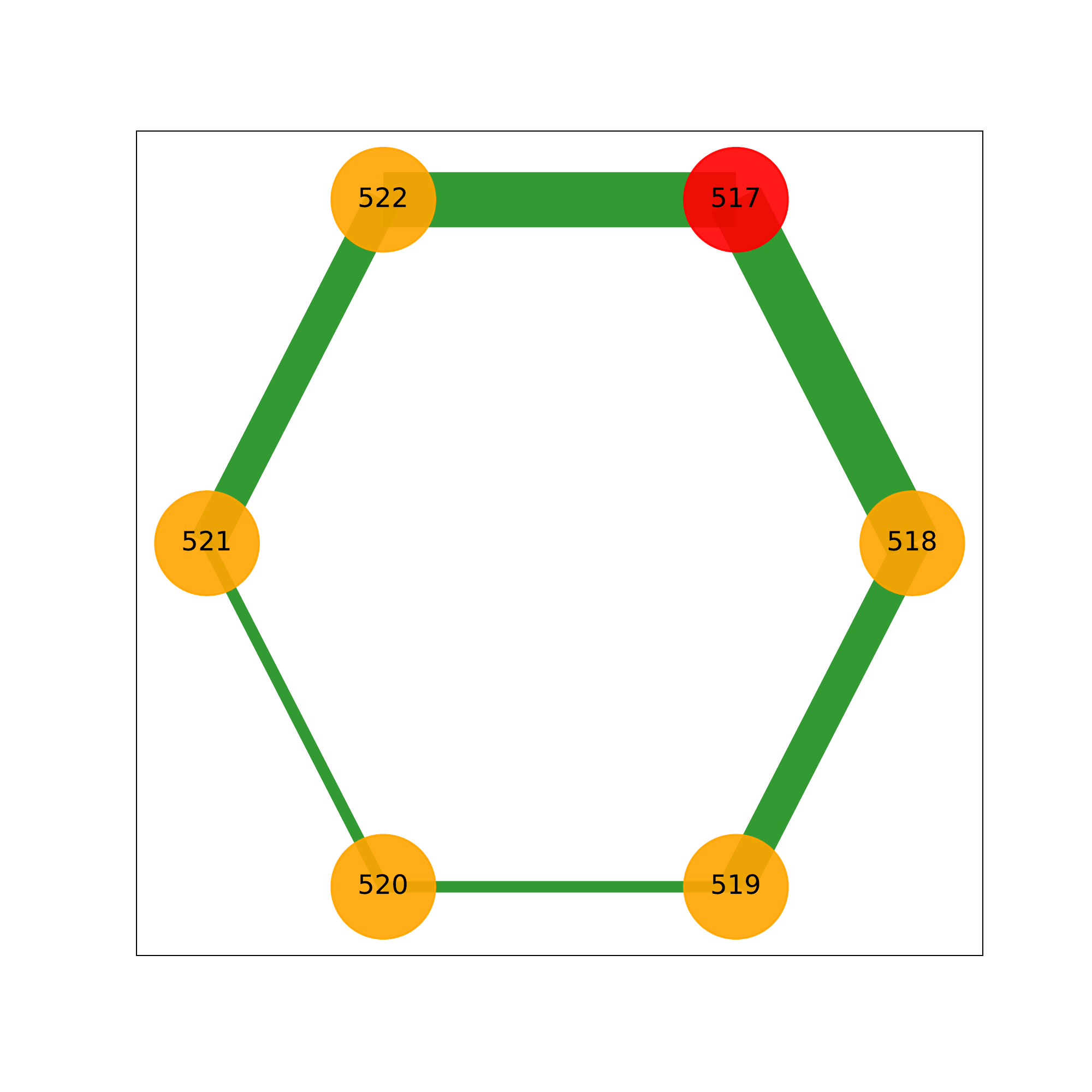}} & 
        \multicolumn{1}{m{2.0cm}}{\includegraphics[clip,trim=2.5cm 2.5cm 2.0cm 2.4cm,width=2.0cm]{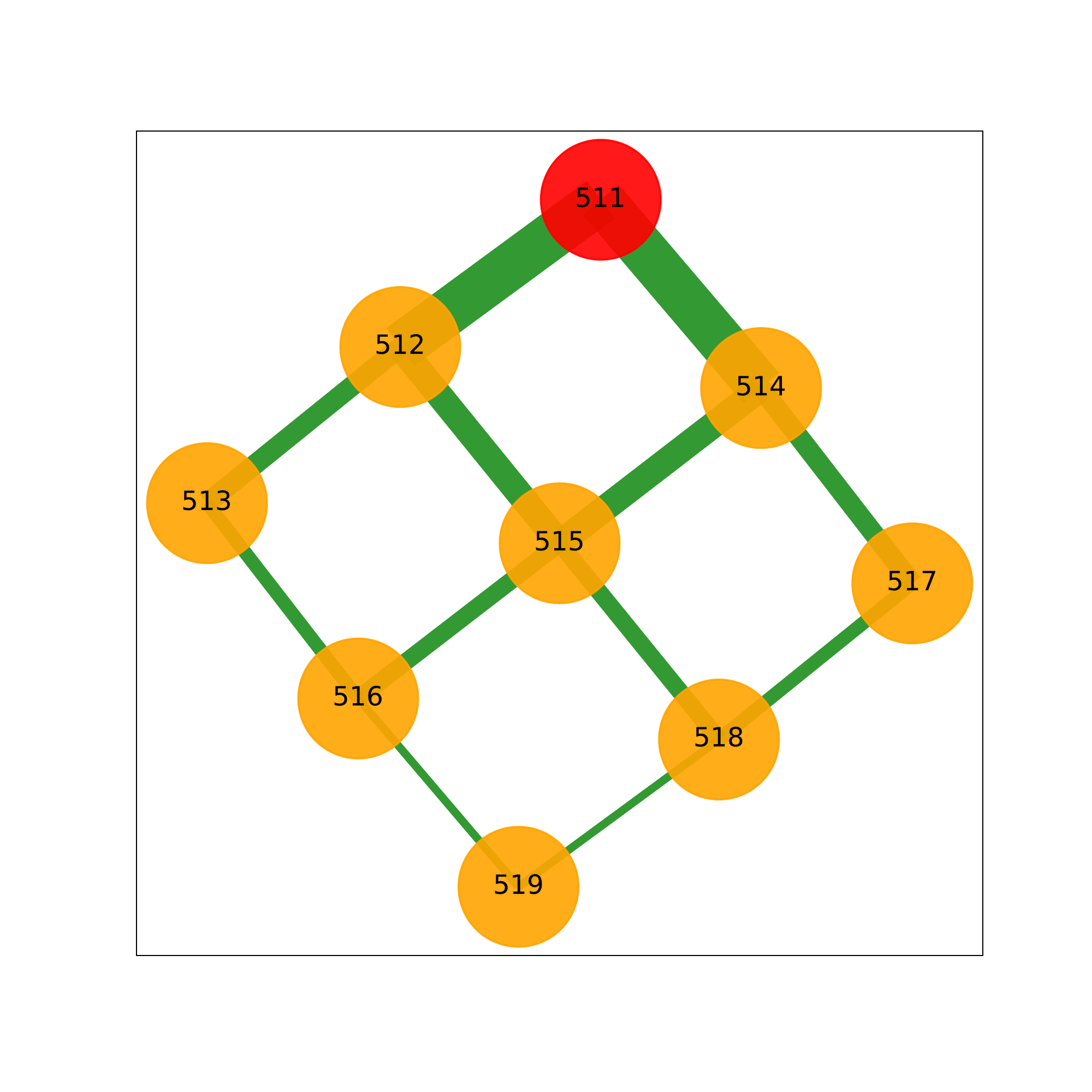}} & 
        \multicolumn{1}{m{2.0cm}}{\includegraphics[clip,trim=2.5cm 2.5cm 2.0cm 2.4cm,width=2.0cm]{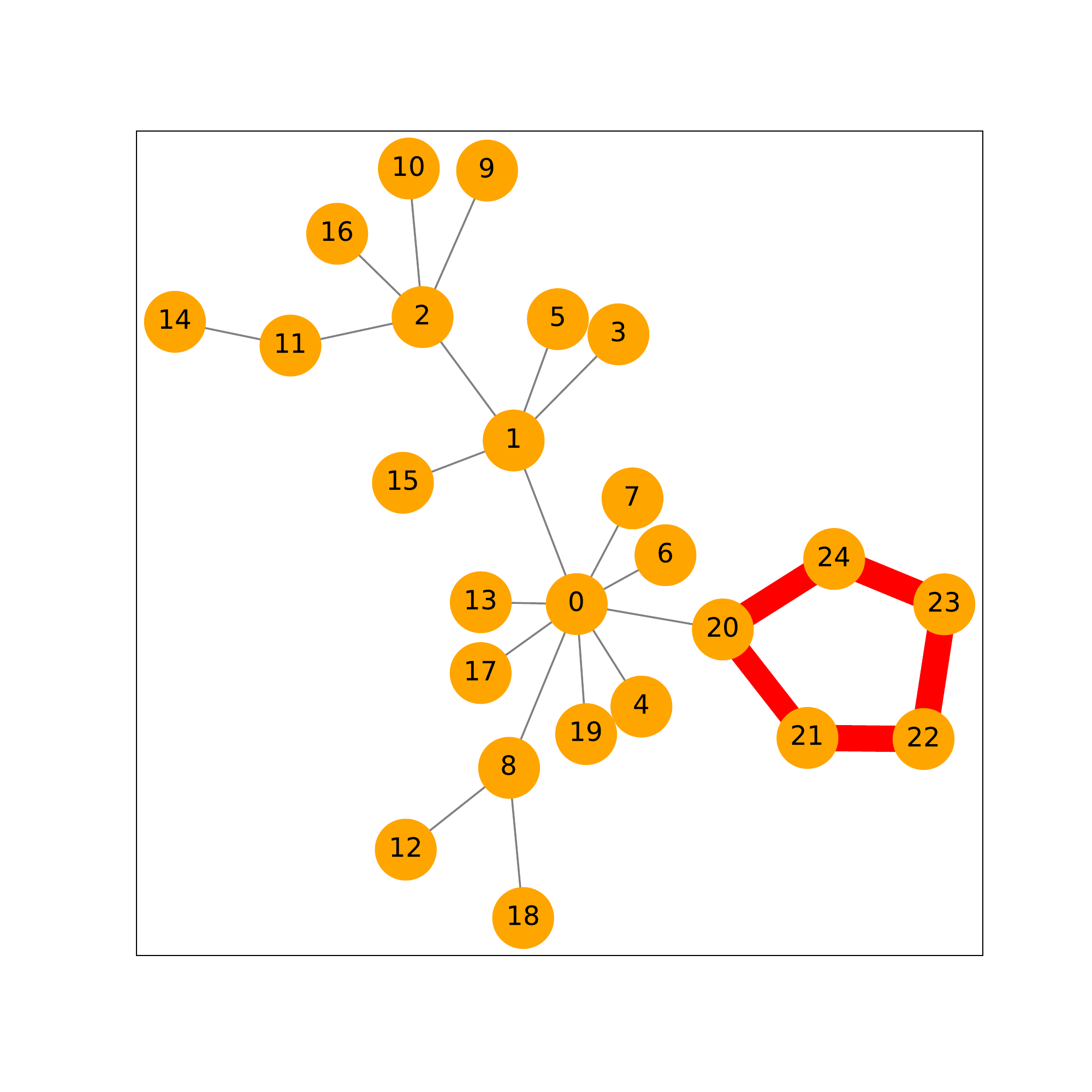}} & 
        \multicolumn{1}{m{2.0cm}}{\includegraphics[clip,trim=2.5cm 2.5cm 2.0cm 2.4cm,width=2.0cm]{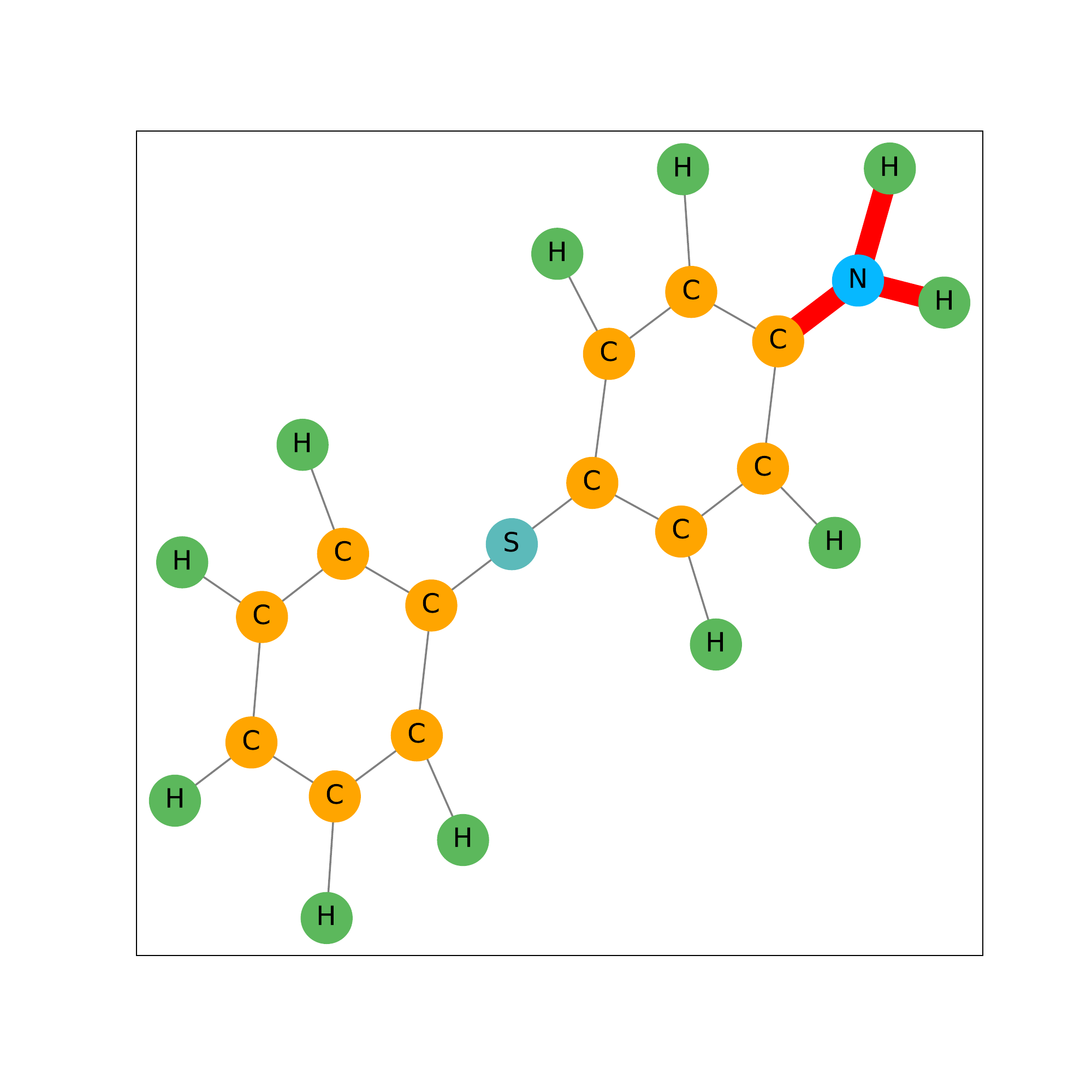}} 
         \\
         & \multicolumn{1}{c}{\scriptsize{(a) BA-Shapes}} & 
        \multicolumn{1}{c}{\scriptsize{(b) BA-Community}} & 
        \multicolumn{1}{c}{\scriptsize{(c) Tree Cycle}} & 
        \multicolumn{1}{c}{\scriptsize{(d) Tree Grid}} & 
        \multicolumn{1}{c}{\scriptsize{(e) BA-2motifs}} & 
         \multicolumn{1}{c}{\scriptsize{(f) Mutag}} 
    \end{tabular}
    
    \caption{Qualitative Comparison between SCALE, PGExplainer, and GNNExplainer. Green/Red edges are the ones selected for explanations. The thicker the edges, the more important they are to predictions. In node classification datasets, only SCALE can highlight the importance of edges.}
    \label{fig:qualitative_comparison}
\end{figure*}

We chose one instance from each dataset and visualized explanations provided by SCALE, GNNExplainer, and PGExplainer in \cref{fig:qualitative_comparison}. SCALE can highlight crucial edges in graph classification explanations similar to two baselines. Since SCALE achieves higher precision scores, its explanations include fewer false positive edges. Even though GNNExplainer and PGExplainer can highlight impactful edges of target nodes in node classification explanations, they cannot differentiate the contributions of these edges since edge weights only represent selection probabilities. Conversely, SCALE visualizes edges with different widths using the probability that a random walker jumps through these edges. This feature makes explanations highly intuitive, wherein the thicker the edges, the more important they are to target nodes. Moreover, we observed that edges started from direct neighbors within the same community received higher walk probabilities. 

\begin{figure}[ht]
    \centering
    \setlength\tabcolsep{3pt}
    \begin{tabular}{r p{2.0cm}p{2.0cm}p{2.0cm}}
        \textbf{\thead{Tree \\ Grid}} &
         \multicolumn{1}{m{2.0cm}}{\includegraphics[clip,trim=2.5cm 2.5cm 2.0cm 2.4cm,width=2.0cm]{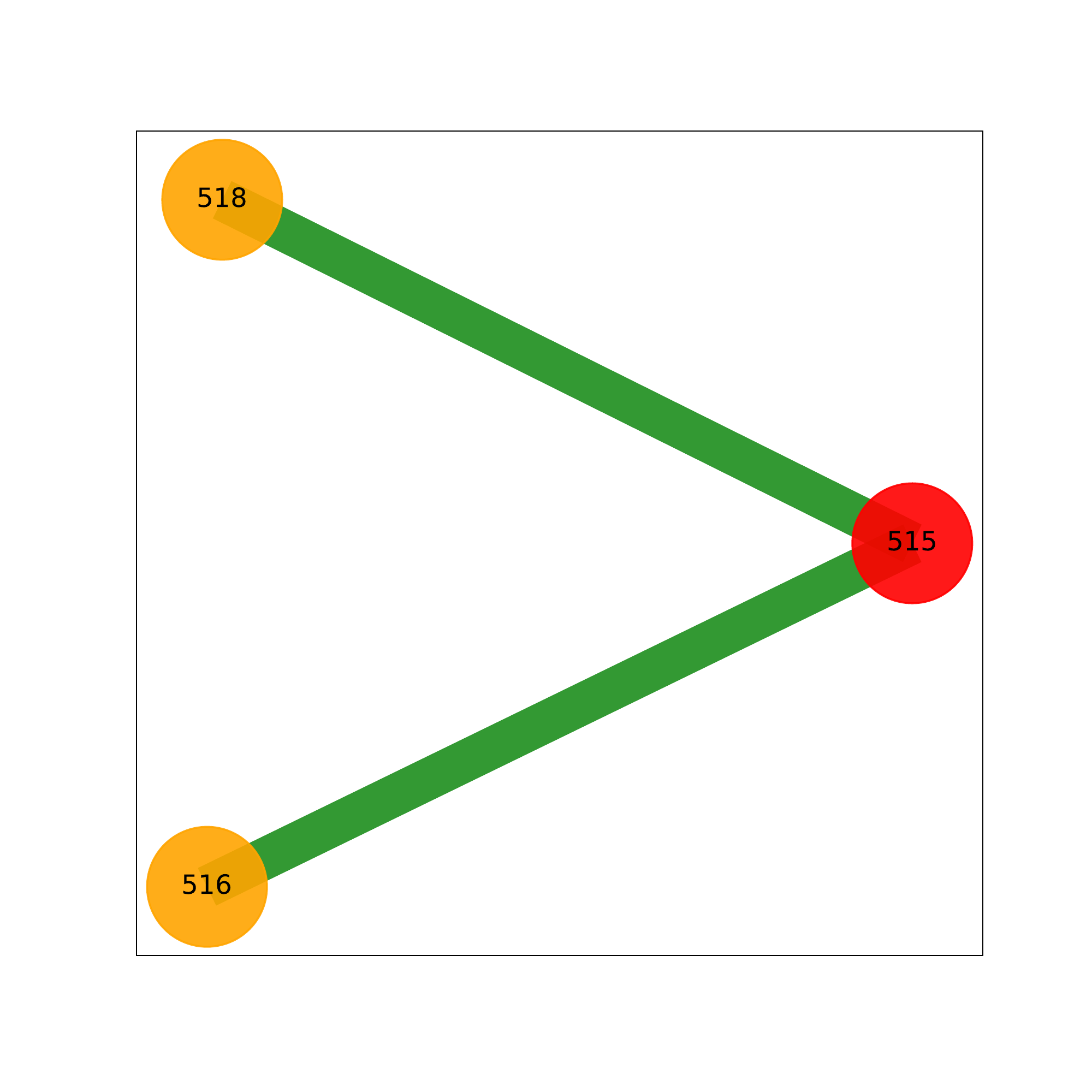} } & 
        \multicolumn{1}{m{2.0cm}}{\includegraphics[clip,trim=2.5cm 2.5cm 2.0cm 2.4cm,width=2.0cm]{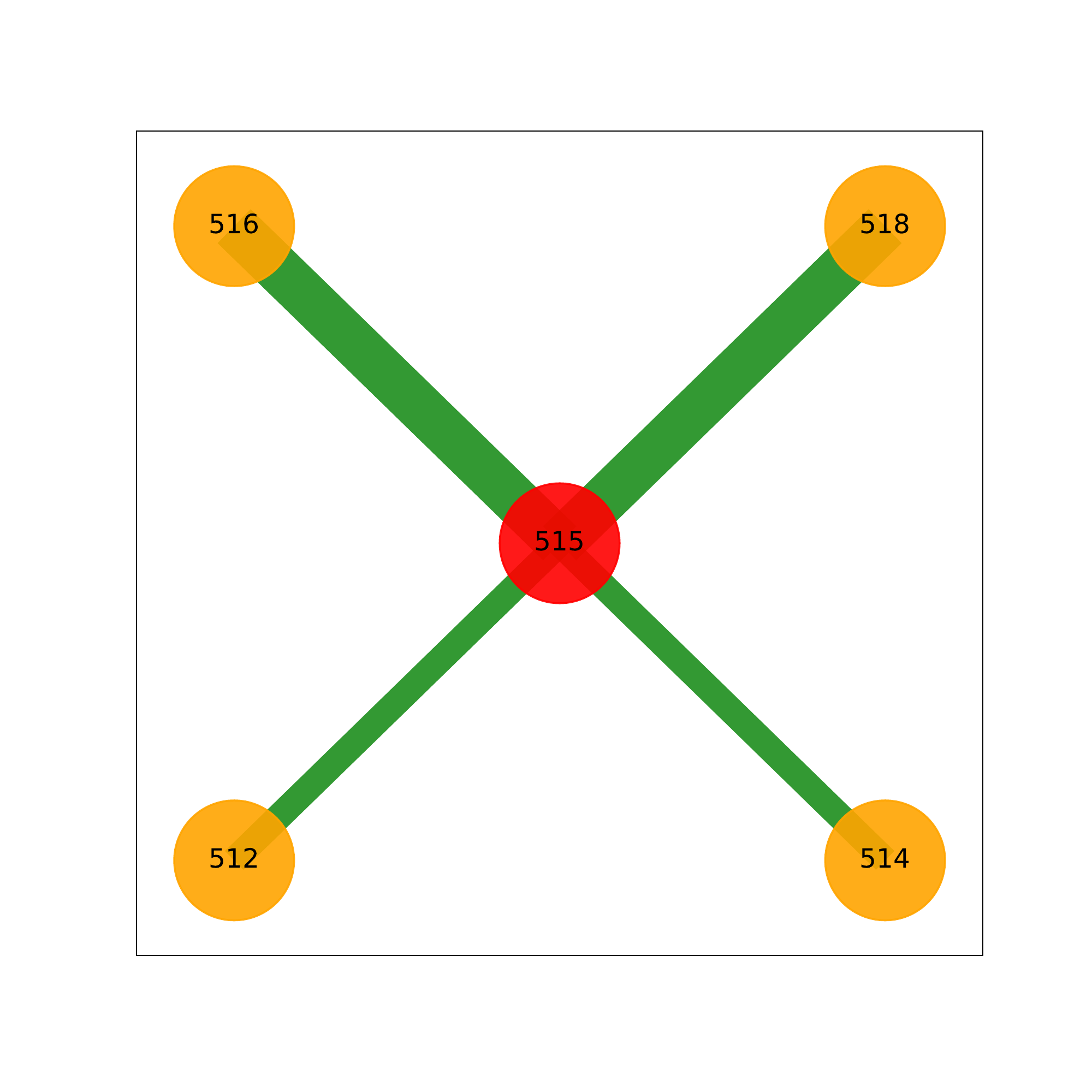} } & 
        \multicolumn{1}{m{2.0cm}}{\includegraphics[clip,trim=2.5cm 2.5cm 2.0cm 2.4cm,width=2.0cm]{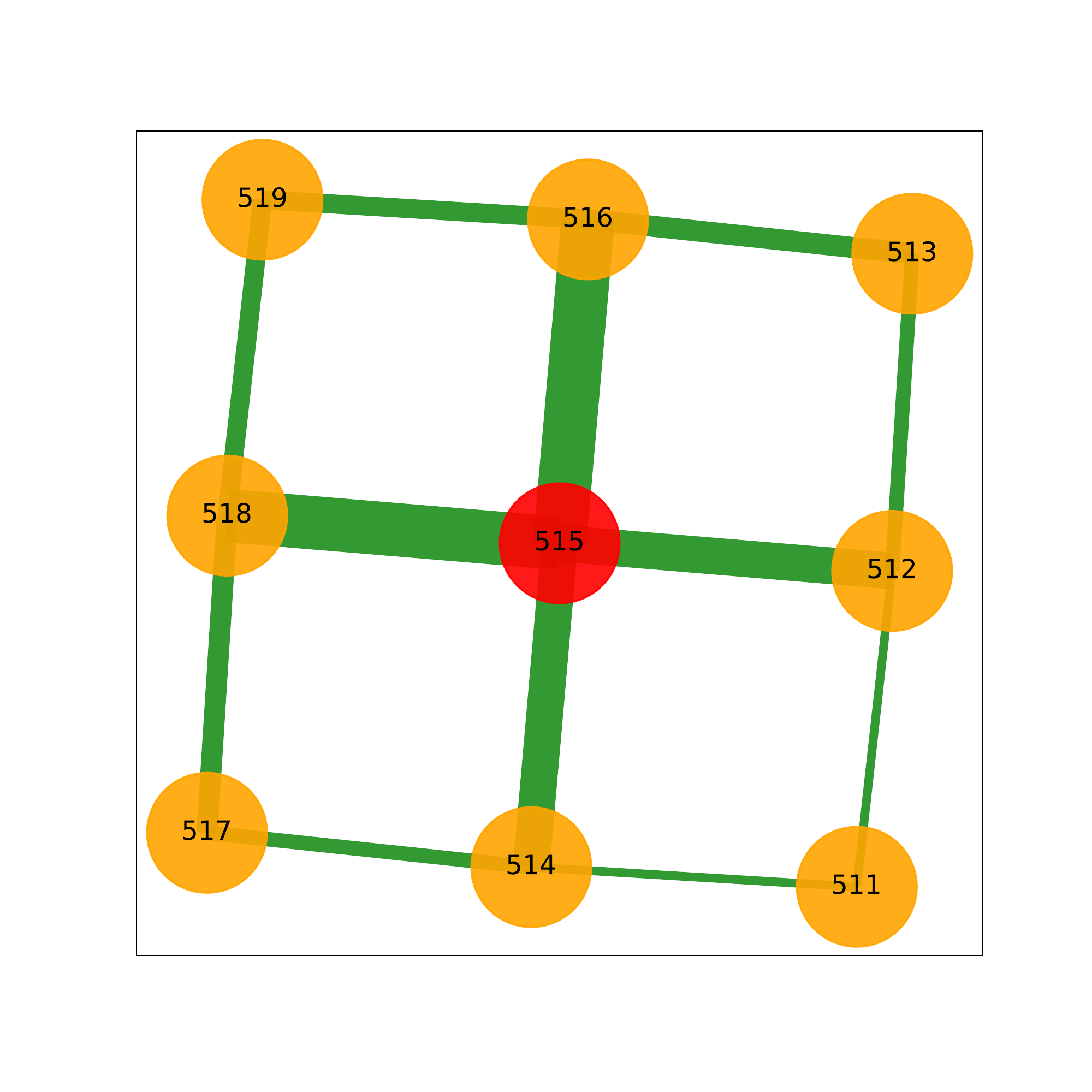} }
         \\
        
        & \multicolumn{1}{c}{\scriptsize{(a) K = 2}} & 
        \multicolumn{1}{c}{\scriptsize{(b) K = 4}} & 
        \multicolumn{1}{c}{\scriptsize{(c) K = 8}} \vspace{0.1cm} \\
        
         \textbf{\thead{Tree \\ Cycle}} & 
         \multicolumn{1}{m{2.0cm}}{\includegraphics[clip,trim=2.5cm 2.5cm 2.0cm 2.4cm,width=2.0cm]{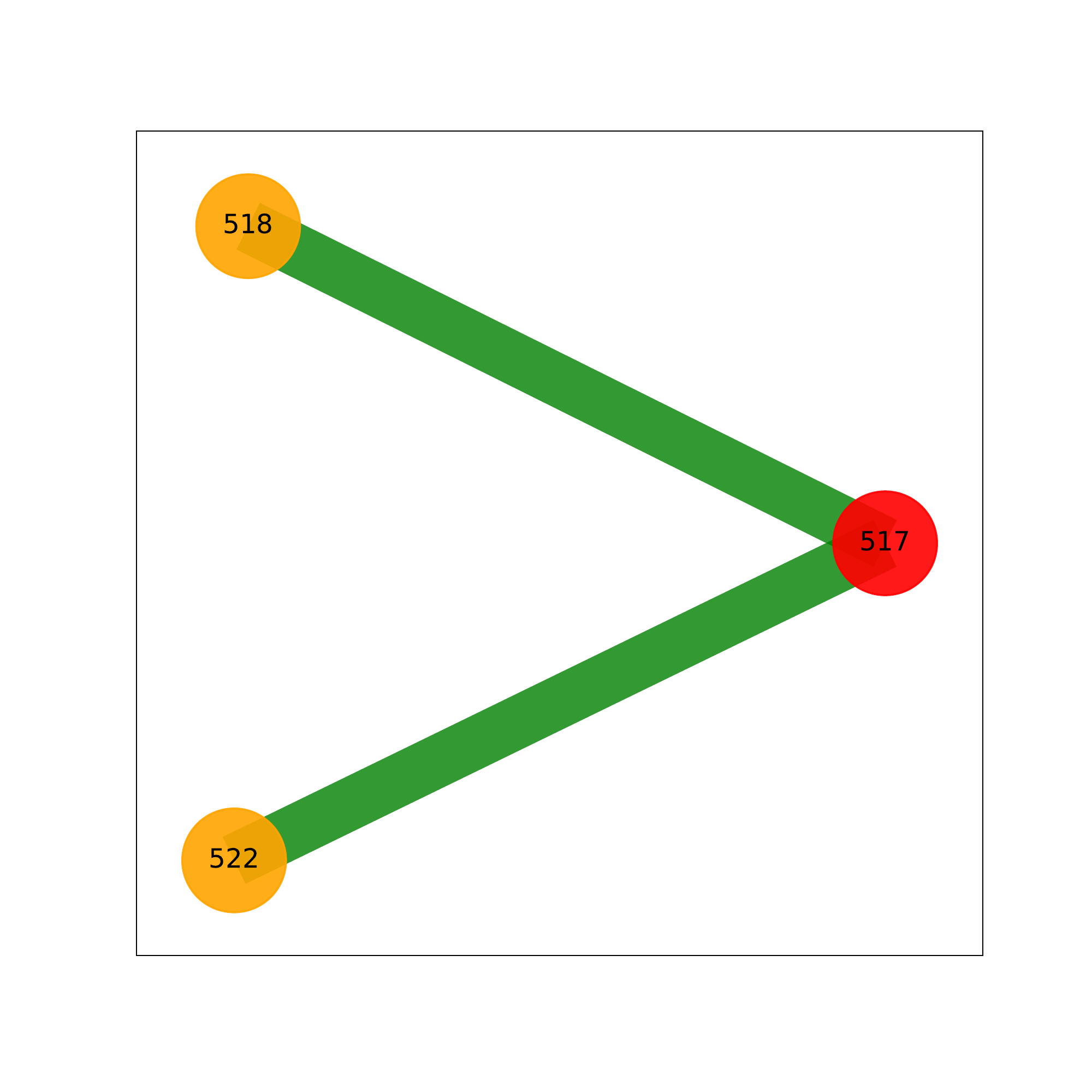}} & 
        \multicolumn{1}{m{2.0cm}}{\includegraphics[clip,trim=2.5cm 2.5cm 2.0cm 2.4cm,width=2.0cm]{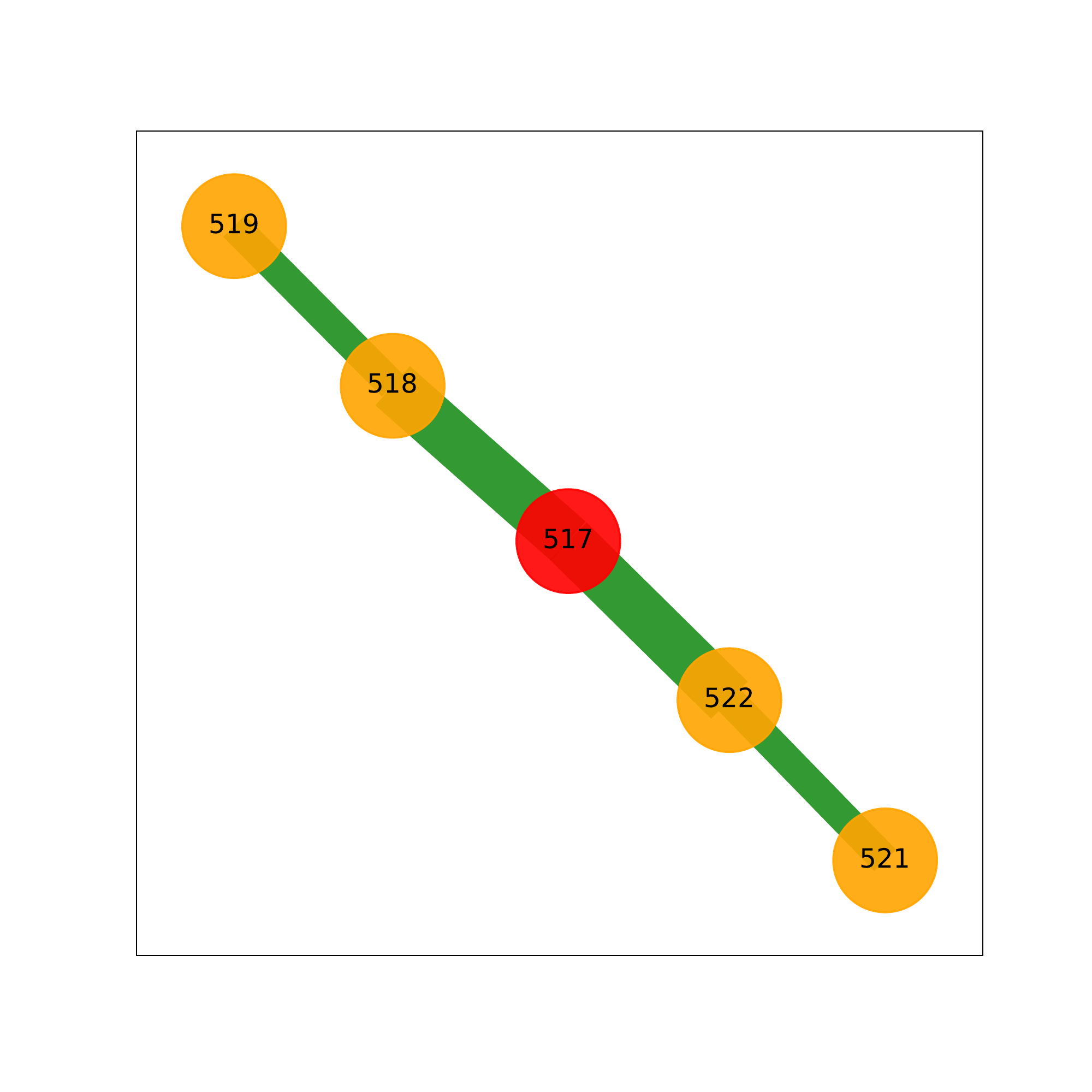}} & 
        \multicolumn{1}{m{2.0cm}}{\includegraphics[clip,trim=2.5cm 2.5cm 2.0cm 2.4cm,width=2.0cm]{figures/graphs/tree_cycle_517_l3.pdf}} 
         \\
         & \multicolumn{1}{c}{\scriptsize{(c) K = 2}} & 
        \multicolumn{1}{c}{\scriptsize{(d) K = 4}} & 
        \multicolumn{1}{c}{\scriptsize{(f) K = 5}} 
    \end{tabular}
    
    \caption{Multi-level Expansion of Structural Explanations for Node-level Predictions. K denotes the number of top influential nodes sorted by importance scores.}
    \label{fig:multi_level}
\end{figure}


Multi-level expansion of explanations is useful in many scenarios, such as recommender systems. Even though GNNExplainer and PGExplainer can also present explanations on multiple levels by adjusting the visibility threshold, setting this value too low may result in outputs with multiple disconnected components due to the independence of edge selections. Edges started from direct (1-hop) neighbors may have lower selection probabilities than those from 2-hop or 3-hop neighbors. Moreover, the maximum number of hops is predefined at the sampling step before training explainable models. Therefore, we cannot expand explanations beyond this limit. Conversely, SCALE's explanations are more intuitive, as presented in \cref{fig:multi_level}. As can be seen, close neighbors receive higher scores than distant ones. Moreover, one can expand explanations by adjusting the visibility threshold or displaying influential neighbors layer by layer without limitations.

\subsection{Efficiency of Explanation Querying Procedure for Node Classification}

Here, we further studied the effectiveness of \cref{querying_rwr} by executing it on adjacency matrices provided by GCN-MLP and GAT. Naively selecting important edges for node classification of target nodes was an inefficient approach, especially in complicated graphs. Therefore, the explanation results given by this approach on the original GCN-MLP and GAT were inaccurate as ground-truth motifs get complicated in Tree-Cycle and Tree-Grid datasets. When applying \cref{querying_rwr} to these models, explanation correctness increased significantly, as presented in the last two rows of \cref{tab:quant_compare}. However, the improved results were still lower than SCALE. Since edge weights represented the influences of neighbors, their correctness highly impacted the RWR procedure and the quality of explanations. Our proposed online KG paradigm enables a self-explainable GNN to capture interactions between nodes efficiently, thus resulting in more accurate edge weights.

\subsection{Evaluating Feature Attribution Module}

%

\begin{figure}[ht]
    \centering
    \includegraphics[width=0.85\columnwidth,trim={0.2cm 0.2cm 0.5cm 1cm},clip]{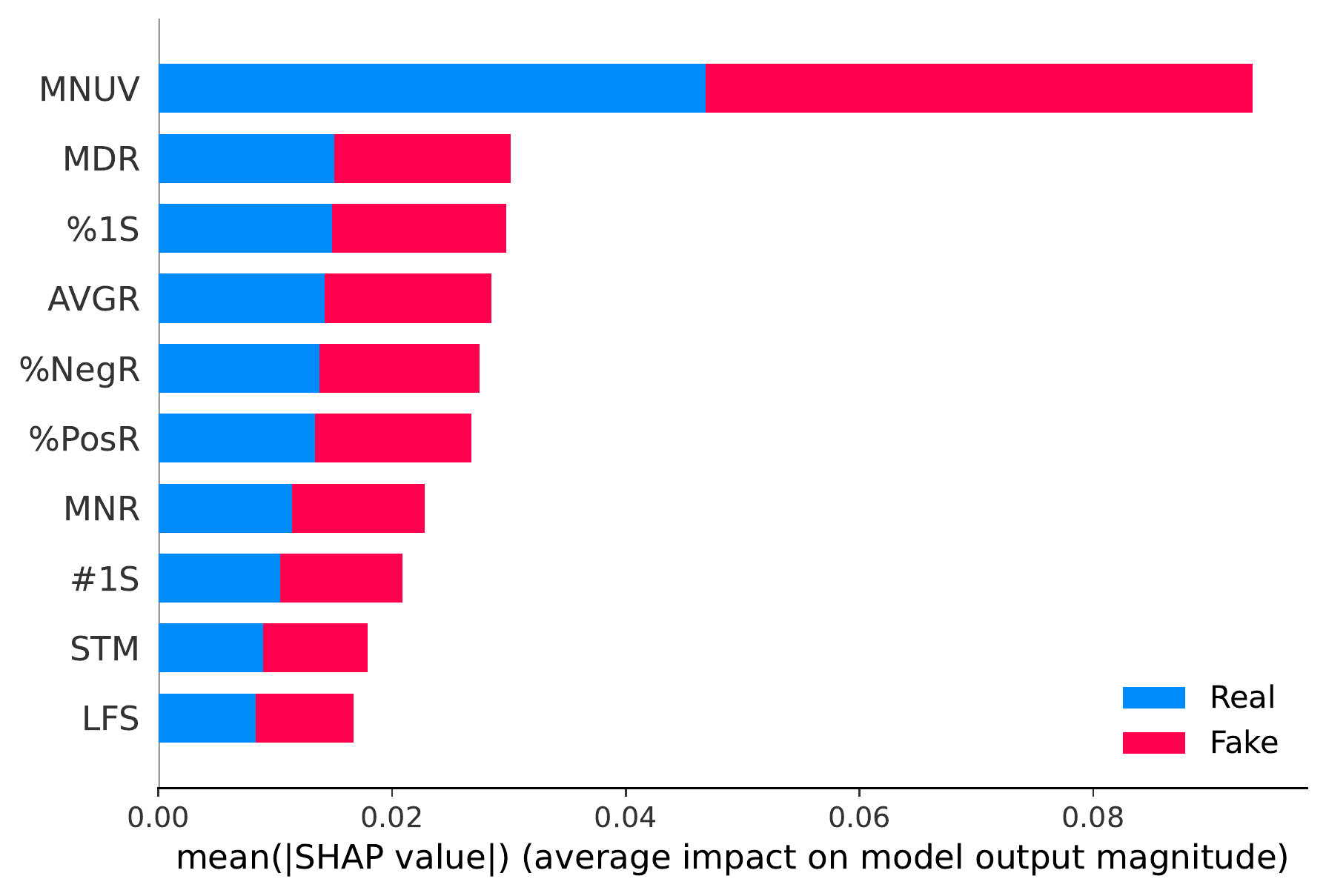}
    \caption{An Overall Summary of Feature Attributions for Node Classification on Amazon Dataset. It summarizes the average impact of features on predictive probabilities. The longer the bar, the more influential a feature is.}
    \label{fig:feature_importance}
\end{figure}

\begin{figure}[ht]
    \centering
    \hfil
    \subfloat[Fraudulent User Class]{
        \includegraphics[width=0.9\columnwidth,trim={2.8cm 0.1cm 0.35cm 0.2cm},clip]{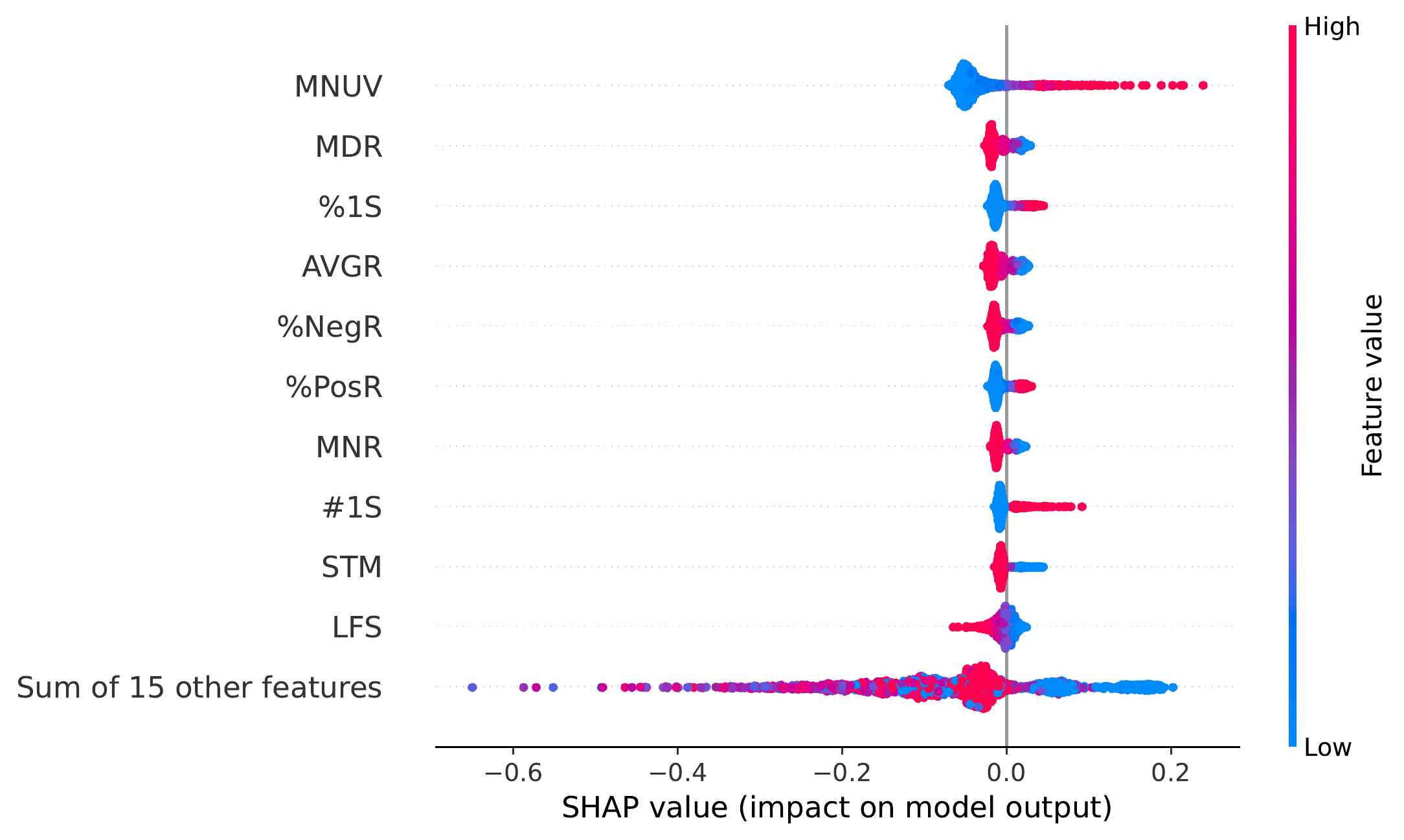}
        \label{fig:feat_a}
    }
    \hfil
    \subfloat[Benign User Class]{
        \includegraphics[width=0.9\columnwidth,trim={2.8cm 0.1cm 0.35cm 0.2cm},clip]{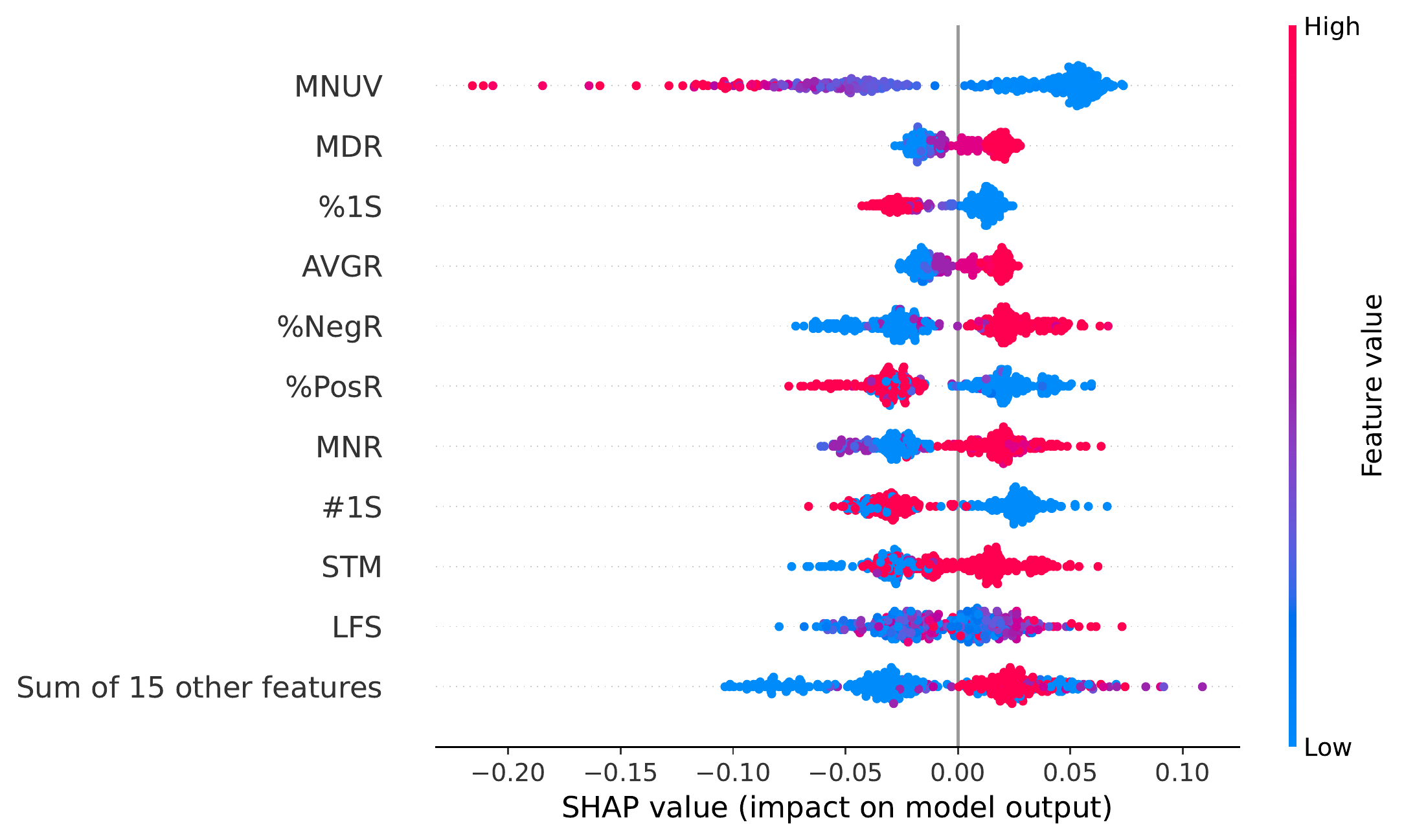}
        \label{fig:feat_b}
    }
    \caption{A Summary of Value-impact Relationships for Each Class in Amazon Dataset. Y-axes are feature names, and X-axes are feature impact values. The color bar in the bottom figures represents the magnitude of values (redder - bigger, bluer - smaller). All figures have the same order of features. }
    \label{fig:feature_importance_2}
\end{figure}

We aimed to evaluate the efficiency of the feature attribution module. The Amazon dataset was used since it contains intelligible node features. Each node in the Amazon graph has 25 statistical properties representing reviewing behaviors of users for products on the Amazon platform. Fraudsters or attackers are users who try to cheat the recommender system to promote particular products. Moreover, they also cover themselves by trying to be like benign users as much as possible. Since ground-truth explanations do not exist, we compared our observations on feature attributions provided by SCALE with insights discovered by Zhang et al. \cite{zhang2020gcn}.

\begin{figure}[ht]
    \centering
    \hfil
    \subfloat[Fraudulent User Class]{
        \includegraphics[width=0.9\columnwidth,trim={0.2cm 3.3cm 2.2cm 0.3cm},clip]{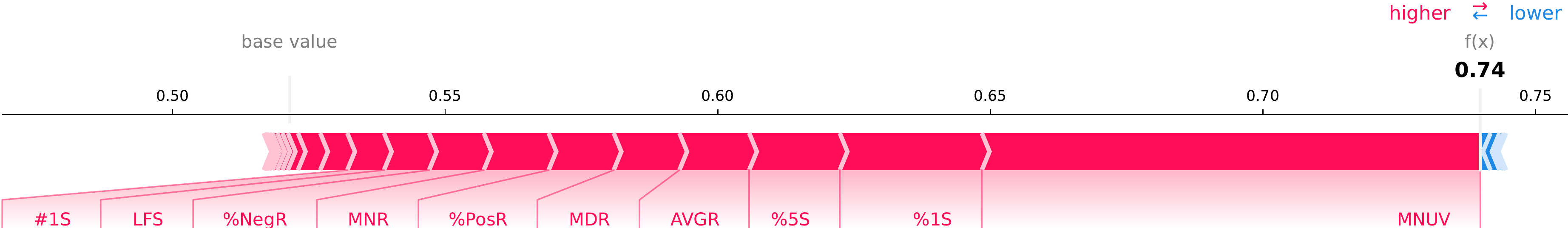}
        \label{fig:example_real}
    }
    \hfil
    \subfloat[Benign User Class]{
        \includegraphics[width=0.9\columnwidth,trim={0.2cm 3.3cm 3.3cm 0.3cm},clip]{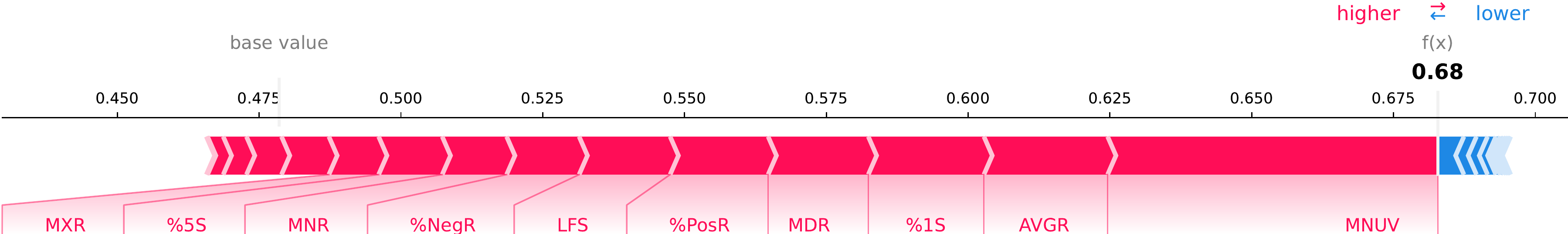}
        \label{fig:example_fake}
    }
    \hfil
    \caption{Examples of Feature Attributions of Predictions for Each User Class. Red color bars present how much each feature increases the predictive probability, while blue color bars demonstrate the impact of features in the opposite direction.}
    \label{fig:instance_explanations}
\end{figure}

Here, overviews of feature contributions and examples of instance-level explanations are reported. Specifically, \cref{fig:feature_importance} presents an overall summary of feature importances, while \cref{fig:feature_importance_2} elaborates on relationships between feature values and their impacts on predictive probabilities. Then, two examples of instance-level explanations for each class are shown in \cref{fig:instance_explanations}. SHAP values in the figures mean the magnitude of marginal probability contributions. Here are descriptions of some features presented in these figures: \textbf{MNUV} - Minimum number of unhelpful votes; \textbf{MDR} - Median of ratings; \textbf{\%1S} - Ratio of 1-star votes; \textbf{AVGR} - Average of ratings; \textbf{\%NegR }- Ratio of negative ratings; \textbf{STM} - Sentiment of feedback; \textbf{LFS} - Length of feedback. Please refer to \cite{rayana2015collective, dou2020enhancing} for the complete description.

Based on the presented reports, we obtained the following observations. The minimum number of unhelpful votes plays a crucial role in model predictions, meaning clear differences exist between the voting patterns of two user classes. As \cref{fig:feature_importance_2} shows, fraudulent users receive numerous negative votes from others since high \textbf{MNVU} values correspond to large probability contributions. We can also see that a large number/ratio of low-star ratings and feedback with negative sentiment increases the possibility that a user is fraudulent. Conversely, regular users give fair ratings and reviews with neutral or positive sentiments. Moreover, fake reviews' feedback summaries are usually shorter than real reviews. These observations match findings from Zhang et al. \cite{zhang2020gcn} that attackers usually give a high rating to a target item (promoted one) and low ratings to other regular items. Therefore, we can say that our proposed method for examining feature contributions is effective and accurate.

\subsection{Ablation Studies}
We conducted a series of ablation studies to study different aspects of SCALE. First, We aimed to select appropriate jumping probability values for particular scenarios. Then, we studied how explanation correctness responds to the changes in model accuracy. Next, different KD settings were evaluated to elaborate on the efficiency of the online KD paradigm. Finally, we studied how distilled knowledge impacts explanations in different classification tasks. 

\pgfplotsset{width=5.05cm,height=4cm}
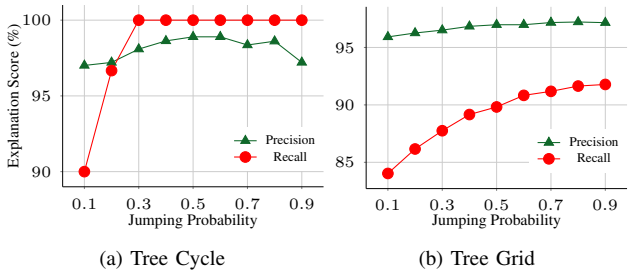
\begin{figure}[ht]
    \centering
    \subfloat[Tree Cycle]{
        \begin{tikzpicture}
            \begin{axis}[
                grid=both,
                grid style={line width=.03pt, draw=gray!40},
                major grid style={line width=.03pt,draw=gray!40},
                mark size=2pt,
                legend style={draw=none,fill=white,at={(1.01,0.33)},anchor=north east,legend columns=1,nodes={scale=0.5, transform shape},column sep=0.08cm},
                x label style={at={(axis description cs:0.5,-0.1)},anchor=north},
                y label style={at={(axis description cs:-0.12,0.5)},anchor=south},
                xtick={0.1,0.3,0.5,0.7,0.9},
                xlabel=Jumping Probability,ylabel=Explanation Score (\%)]
                \addplot[color={rgb,255:red,16;green,104;blue,42},mark=triangle*] coordinates{
                    (0.1,97.01)
                    (0.2,97.21)
                    (0.3,98.09)
                    (0.4,98.63)
                    (0.5,98.9)
                    (0.6,98.9)
                    (0.7,98.36)
                    (0.8,98.61)
                    (0.9,97.21)
                        };
                \addlegendentry{Precision}
        
                \addplot[color={red},mark=*] coordinates{
                    (0.1,90.0)
                    (0.2,96.67)
                    (0.3,100)
                    (0.4,100)
                    (0.5,100)
                    (0.6,100)
                    (0.7,100)
                    (0.8,100)
                    (0.9,100)
                        };
                \addlegendentry{Recall}
            \end{axis}
            \end{tikzpicture}
    }
    \hfil
    \hspace{-0.5cm}
    \subfloat[Tree Grid]{
        \begin{tikzpicture}
            \begin{axis}[
                grid=both,
                grid style={line width=.03pt, draw=gray!40},
                major grid style={line width=.03pt,draw=gray!40},
                mark size=2pt,
                legend style={draw=none,fill=white,at={(1.01,0.33)},anchor=north east,legend columns=1,nodes={scale=0.5, transform shape},column sep=0.08cm},
                x label style={at={(axis description cs:0.5,-0.09)},anchor=north},
                y label style={at={(axis description cs:-0.09,0.5)},anchor=south},
                xtick={0.1,0.3,0.5,0.7,0.9},
                xlabel=Jumping Probability,ylabel=\empty]
                \addplot[color={rgb,255:red,16;green,104;blue,42},mark=triangle*] coordinates{
                    (0.1,95.92)
                    (0.2,96.26)
                    (0.3,96.51)
                    (0.4,96.84)
                    (0.5,96.98)
                    (0.6,96.98)
                    (0.7,97.17)
                    (0.8,97.22)
                    (0.9,97.16)
                        };
                \addlegendentry{Precision}
        
                \addplot[color={red},mark=*] coordinates{
                    (0.1,84.03)
                    (0.2,86.16)
                    (0.3,87.75)
                    (0.4,89.16)
                    (0.5,89.82)
                    (0.6,90.83)
                    (0.7,91.18)
                    (0.8,91.64)
                    (0.9,91.78)
                        };
                \addlegendentry{Recall}
            \end{axis}
            \end{tikzpicture}
    }
    \hfil
    
\caption{Relationships Between Jumping Probability and Explanation Scores. A reasonable probability should be between 0.5 and 0.9.}
\label{fig:jumping_scores}
\end{figure}
\noindent\textbf{Relationships Between Jumping Probability and Explanation Correctness.} We have the following observations based on experiments conducted on Tree-Cycle and Tree-Grid datasets. A random walker tends to restart more often with a small probability, whereas it explores new states with a large value. As illustrated in \cref{fig:jumping_scores}, small jumping probabilities cause low precision and recall scores, especially when many hops are required to complete the motifs. In the Tree-Cycle dataset, the precision score gradually improves as the jumping probability increases to 0.6 but decreases when the probability exceeds this value. In the Tree-Grid dataset, precision and recall scores correspond to the magnitude of the jumping probability since the complexity of grid motifs requires long walks to traverse all nodes in ground-truth explanations. Therefore, the large probability is appropriate for the Tree-Grid case, while the probability between 0.5 and 0.6 is better for the Tree-Cycle dataset. In practice, a reasonable value can range between 0.5 and 0.9 depending on the characteristics of graphs in particular scenarios.

\pgfplotsset{width=7cm,height=4.5cm}
\begin{figure}[ht]
    \centering
    \begin{tikzpicture}
    \begin{axis}[
        grid=both,
        grid style={line width=.03pt, draw=gray!40},
        major grid style={line width=.03pt,draw=gray!40},
        mark size=2pt,
        legend style={draw=none,fill=white,at={(1.01,0.83)},anchor=north east,legend columns=1,nodes={scale=0.5, transform shape},column sep=0.08cm},
        x label style={at={(axis description cs:0.5,-0.09)},anchor=north},
        y label style={at={(axis description cs:-0.09,0.5)},anchor=south},
        xtick={70,80,90},
        xlabel=Model Accuracy (\%),ylabel=Explanation Score (\%)]
        \addplot[color={rgb,255:red,16;green,104;blue,42},mark=triangle*] coordinates{
            (70.000000,52.03)
            (80.000000,64.41)
            (90.000000,66.18054054)
                };
        \addlegendentry{Precision}

        \addplot[color={red},mark=*] coordinates{
            (70.000000,99.50)
            (80.000000,99.64)
            (90.000000,99.72)
                };
        \addlegendentry{Recall}
    \end{axis}
    \end{tikzpicture}
\caption{Model Accuracy and Explanation Correctness Relationships on Mutag Dataset. The explanation correctness is proportional to the model accuracy.}
\label{fig:model_explanation_score}
\end{figure}
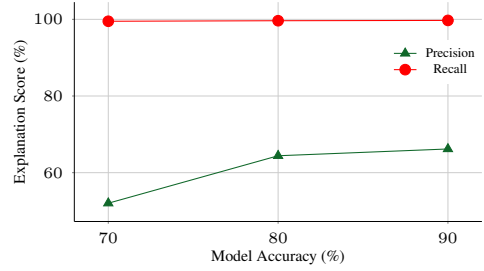
\noindent\textbf{Relationships Between Model Accuracy and Explanation Correctness.}
Most existing GNN explanation methods assume that pre-trained models are extremely accurate. Here, we studied how the explanation correctness is impacted when the model accuracy decreases in the Mutag dataset. As \cref{fig:model_explanation_score} depicts, the precision score increases significantly as the model accuracy improves from 70\% to 80\%, meaning fewer false positive edges are included in explanations. When the precision accuracy increases to 90\%, the precision score only improves slightly, suggesting that 80\% is an acceptable value for extracting influential subgraphs in this dataset. We can conclude that explanations are more relevant and accurate as the model accuracy increases.

\noindent\textbf{How effective is Online KD?}
This experiment specifies the effect of distilled knowledge, including embedding vectors and predictive distributions taken from the black-box GNN, on the correctness of structural explanations. We used the Mutag dataset for this experiment. We compared the experimental results of four settings on the self-explainable GNN as follows:
\begin{itemize}
    \item \textbf{Naive}: The learner uses neither embedding vectors nor predictive distributions in training.
    \item \textbf{Embed}: The learner uses only embedding vectors to initialize learnable masks and sets $\lambda = 0$ in $\mathcal{L}^s$.
    \item \textbf{KDL}: The learner does not use embedding vectors to initialize learnable masks.
    \item \textbf{Joint}: The learner uses both components in training.
\end{itemize}
\pgfplotsset{width=4.9cm,height=4cm}
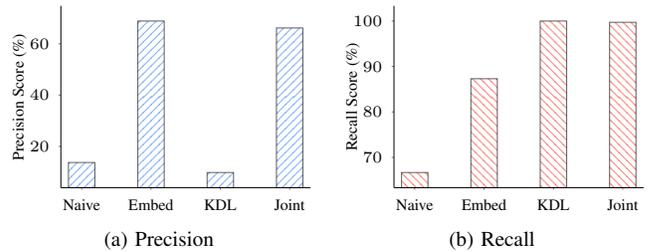
\begin{figure}[ht]
    \centering
    \subfloat[Precision]{
        \begin{tikzpicture}
            \begin{axis}[
                ybar,
                legend cell align=left,
                area legend,
                bar width=10pt,
                ylabel={Precision Score (\%)},
                y label style={at={(axis description cs:-0.1,0.5)},anchor=south},
                symbolic x coords={Naive, Embed, KDL, Joint},
                ]
                \addplot[pattern color={rgb,255:red,66;green,133;blue,244},draw opacity=0.6,pattern=north east lines] coordinates{(Naive, 13.71) (Embed, 68.89) (KDL, 9.80) (Joint, 66.18)};
            \legend{}
            \end{axis}
        \end{tikzpicture}
    }
    \hfill
    \subfloat[Recall]{
        \begin{tikzpicture}
            \begin{axis}[
                ybar,
                legend cell align=left,
                area legend,
                bar width=10pt,
                ylabel={Recall Score (\%)},
                y label style={at={(axis description cs:-0.1,0.5)},anchor=south},
                symbolic x coords={Naive, Embed, KDL, Joint},
                ]
                \addplot[pattern color={rgb,255:red,254;green,67;blue,53},draw opacity=0.6,pattern=north west lines] coordinates{(Naive, 66.67) (Embed, 87.31) (KDL, 100) (Joint, 99.72)};
            \legend{}
            \end{axis}
        \end{tikzpicture}
    }
    \hfill
    \caption{Explanation Correctness in Different Settings of the Online Knowledge Distillation Paradigm. SCALE provides the most accurate explanations when initializing a learnable mask with a black-box model's embedding vectors and using the KD loss for training learners.}
    \label{fig:online_kg_ab}
\end{figure}

Each setting was executed five times and reported the average result. As presented in \cref{fig:online_kg_ab}, precision and recall scores are high in the \textbf{Joint} case but significantly low in the \textbf{Naive} case. Furthermore, the explanation model achieved good precision in the \textbf{Embed} scenario, which was even better than the \textbf{Joint} case. In this case, many true positive edges were not selected in explanations causing high precision and low recall. Conversely, the learner could not learn to eliminate unimportant edges in the \textbf{KDL} setting, thus resulting in high recall and low precision scores. These results demonstrated that embedding vectors from the black-box GNN are essential for mask initialization. Moreover, distilled information from the predictive distribution of the black-box GNN enhanced the learning process of the self-explainable GNN. 

\noindent\textbf{Studying the Balancing Factor.} Here, we aimed to assess the effect of the balancing factor $\lambda$ in the joint objective function of student models on explanation correctness. We conducted experiments with all datasets and selected results from the Tree-Cycle and BA-2motifs datasets to present in \cref{fig:balancing_factor} as variations in precision/recall scores were evident for them. 

\pgfplotsset{width=9cm,height=4.1cm}
\begin{figure}[htbp]
    \centering
    \subfloat[Precision]{
        \begin{tikzpicture}
        \begin{axis}[
            x=1cm,
            ybar,
            legend cell align=left,
            area legend,
            bar width=6pt,
            ylabel={Precision Score (\%)},
            y label style={at={(axis description cs:-0.06,0.5)},anchor=south},
            symbolic x coords={$\lambda$ =  0.1,$\lambda$ =  0.3,$\lambda$ =  0.5,$\lambda$ =  1,$\lambda$ =  2,$\lambda$ =  4,$\lambda$ =  10},
            legend style={font=\scriptsize, at={(0.5,1.02)}, anchor=south,legend columns=-1,/tikz/every even column/.append style={column sep=0.5cm}},
            ]
            \addplot[pattern color={rgb,255:red,66;green,133;blue,244},draw opacity=0.6,pattern=north east lines] coordinates{($\lambda$ =  0.1, 100) ($\lambda$ =  0.3, 100) ($\lambda$ =  0.5, 98.89) ($\lambda$ =  1, 88.89) ($\lambda$ =  2, 93.33) ($\lambda$ =  4, 94.34) ($\lambda$ =  10, 88.89)};
            \addplot[pattern color=red,draw opacity=0.6,pattern=north west lines] coordinates{($\lambda$ =  0.1, 46.6) ($\lambda$ =  0.3, 61.9) ($\lambda$ =  0.5, 47.37) ($\lambda$ =  1, 40.9) ($\lambda$ =  2, 85.73) ($\lambda$ =  4, 100) ($\lambda$ =  10, 100)};
        
        \legend{Tree Cycle,BA-2motifs}
        \end{axis}
        \end{tikzpicture}
    }
    \hfil
    \vspace{-0.3cm}
    \subfloat[Recall]{
        \begin{tikzpicture}
        \begin{axis}[
            x=1cm,
            ybar,
            legend cell align=left,
            area legend,
            bar width=6pt,
            ylabel={Recall Score (\%)},
            y label style={at={(axis description cs:-0.06,0.5)},anchor=south},
            symbolic x coords={$\lambda$ =  0.1,$\lambda$ =  0.3,$\lambda$ =  0.5,$\lambda$ =  1,$\lambda$ =  2,$\lambda$ =  4,$\lambda$ =  10}
            ]
            \addplot[pattern color={rgb,255:red,66;green,133;blue,244},draw opacity=0.6,pattern=north east lines] coordinates{($\lambda$ =  0.1, 100) ($\lambda$ =  0.3, 100) ($\lambda$ =  0.5, 98.89) ($\lambda$ =  1, 88.89) ($\lambda$ =  2, 93.33) ($\lambda$ =  4, 94.34) ($\lambda$ =  10, 88.89)};
            \addplot[pattern color=red,draw opacity=0.6,pattern=north west lines] coordinates{($\lambda$ =  0.1, 46.6) ($\lambda$ =  0.3, 61.19) ($\lambda$ =  0.5, 47.37) ($\lambda$ =  1, 40.9) ($\lambda$ =  2, 85.73) ($\lambda$ =  4, 100) ($\lambda$ =  10, 100)};
        \legend{}
        \end{axis}
        \end{tikzpicture}
    }
    \caption{Impact of Balancing Factor $\lambda$ on Explanation Correctness. SCALE obtains accurate explanations with $\lambda \geq 2$ in graph classification problems and $\lambda \leq 1$ in node classification problems.}
    \label{fig:balancing_factor}
\end{figure}
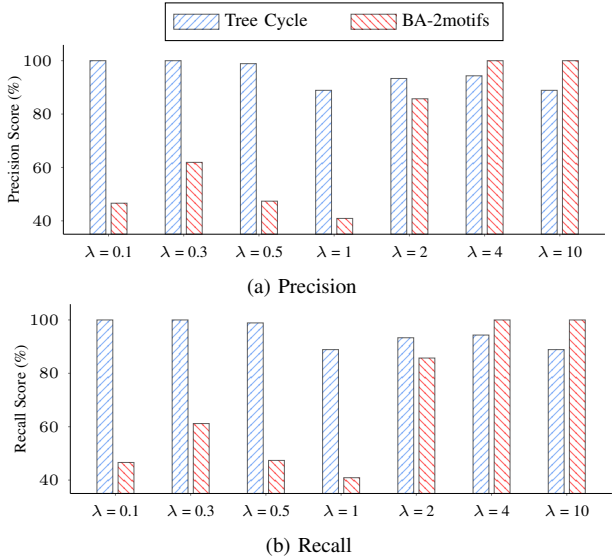

For graph classification datasets, the distilled knowledge from a black-box GNN plays a crucial role in training a self-explainable GNN model. Specifically, SCALE got better results with $\lambda \geq 2$. Even though the self-explainable model obtained high prediction accuracy, it could not provide accurate explanations when $\lambda \leq 1$. Furthermore, explanation correctness also slightly decreased when $\lambda \geq 4$, which means that too much information from the black box model is also not helpful for the explainable model.

Unlike graph classification, $\lambda$ values smaller than one allowed SCALE to provide accurate explanations for node classification problems. The explanation correctness gradually increased as the balancing factor decreased when $\lambda \le 1$. These results suggested that student models cannot deal with too much knowledge distilled from the teacher model. 

\section{Discussion} \label{discussion}
\subsection{Limitations and Future Improvements}
Even though SCALE has several advantages over existing methods, it also contains some limitations. First, it does not consider the explanation problem in link prediction. This problem can be solved using the same approach as explaining node-level predictions. Second, learners and the black-box GNN model are trained in a single thread. In practice, we can accelerate training processes further via distributed settings. Third, the RWR algorithm is executed in a naive way in experiments that will suffer from long execution time for enormous real-world graphs. Since RWR is widely adopted in many large-scale systems, several techniques, such as \cite{tong2006fast}, can be applied to accelerate this algorithm. Fourth, structural explanations for graph classification only show selected influential edges without detailed importance scores. We can further improve the quality of explanations by implementing game theoretic methods like \cite{lundberg2017unified} to compute exact edge contributions since we only need to consider combinations of crucial edges. Next, this paper does not study interactions among graph structures and node features due to their complexity. Methods such as \cite{tsang2020does,lpa_gcn} can be integrated into SCALE to increase its explanation capability. Finally, as the number of learners increases with different explanation aspects, combining their outputs is a promising approach that allows the explanation framework to match users' preferences.

\subsection{Applications} 
SCALE can be applied to numerous applications and systems since it can instantly provide accurate explanations. For instance, recommender systems \cite{wu2020graph,fan2019graph} can easily integrate SCALE into their engines to provide explanation functions to increase the transparency of their systems. SCALE can also provide valuable insights and accelerate research in several domains, such as human action recognition \cite{sun2022human}, bioinformatics \cite{zhang2021graph}, and to name a few. Moreover, feature attributions are beneficial in scenarios when graphs contain intelligible node/edge features. Therefore, SCALE is useful for easing the difficulties in finding insights from input features in graph analytics problems.

\section{Conclusion} \label{conclusion_part}
This paper presented SCALE, the first explanation framework that trained multiple specialty learners to explain GNNs since it was complex to construct one single explainer to examine attributions of factors in an input graph. We aimed to formulate explanation problems as general as post-hoc GNN explanation methods and achieve the explanation speed of self-explainable models. SCALE concentrated on identifying influential factors affecting model predictions from graph structures and node features. It provided explanations with more detailed information compared to existing methods. To achieve these goals, it trained specialty learners simultaneously with a black-box GNN model based on online knowledge distillation. After training, predictions and explanations were provided instantly, wherein several explainers examined different contributions of factors. Specifically, edge masking and random walk with restart procedures were implemented to provide structural explanations for graph-level and node-level predictions. Node feature attributions at different levels were provided by executing a fast feature attribution method on top of a trained MLP learner. Extensive experiments and ablation studies demonstrated SCALE's capabilities and performance superiority. 

\bibliographystyle{IEEEtran}
\bibliography{bibliography}

\end{document}